%% file: main.tex
\documentclass[letterpaper, 10pt, conference]{ieeeconf}

\IEEEoverridecommandlockouts

\usepackage[USenglish]{babel}

\usepackage{mathtools}
\usepackage{amsmath}
\usepackage{amssymb}
\usepackage{amsfonts}
\usepackage{dsfont}
\usepackage{bm}
\usepackage{physics}
\usepackage[separate-uncertainty]{siunitx}

\usepackage{threeparttable}
\usepackage[ruled, vlined, linesnumbered]{algorithm2e}
\usepackage{microtype}
\usepackage[acronym, toc]{glossaries}
\usepackage{graphicx}
\usepackage[caption=false, font=footnotesize]{subfig}  
\usepackage{booktabs}
\usepackage{multirow}
\usepackage{url}
\usepackage{setspace}

\newcommand{\qed}{\hfill\ensuremath{\square}}

\usepackage{xcolor}

\PassOptionsToPackage{table}{xcolor}
\usepackage{graphicx}
\usepackage[table]{xcolor} 
\definecolor{lightgray}{gray}{0.9}

\glsdisablehyper

\SetCommentSty{mycommfont}

\newacronym{TAMP}{TAMP}{Task and Motion Planning}
\newacronym{SVGD}{SVGD}{Stein variational gradient descent}
\newacronym{DSVGD}{DSVGD}{SVGD for discrete distributions}
\newacronym{DMP}{DMP}{dynamic motion primitives}
\newacronym{SGD}{SGD}{stochastic gradient descent}
\newacronym{LGP}{LGP}{logic-geometric programming}
\newacronym{KL}{KL}{Kullback-Leibler}
\newacronym{STAMP}{STAMP}{Stein Task and Motion Planning}

\DeclareMathOperator{\softmax}{softmax}

\usepackage{hyperref}
\hypersetup{
    colorlinks=true,      
    linkcolor=blue,        
    citecolor=green,      
    filecolor=magenta,    
    urlcolor=blue         
}

\title{\LARGE \bf
STAMP: Differentiable Task and Motion Planning \\ via Stein Variational Gradient Descent
}

\author{
    Yewon~Lee\textsuperscript{1},
    Andrew~Z.~Li\textsuperscript{1},
    Philip~Huang\textsuperscript{2},
    Eric~Heiden\textsuperscript{3},
    Krishna~Murthy~Jatavallabhula\textsuperscript{4},\\
    Fabian~Damken\textsuperscript{1,5},
    Kevin~Smith\textsuperscript{4},
    Derek~Nowrouzezahrai\textsuperscript{6},        
    Fabio~Ramos\textsuperscript{3,7},
    Florian~Shkurti\textsuperscript{1}
    \thanks{
        \textsuperscript{1}University of Toronto,
        \textsuperscript{2}Carnegie Mellon University,
        \textsuperscript{3}NVIDIA,
        \textsuperscript{4}Massachusetts Institute of Technology,
        \textsuperscript{5}Technical University of Darmstadt,
        \textsuperscript{6}McGill University,
        \textsuperscript{7}University of Sydney. 
        Corresponding author: Yewon Lee (\texttt{yewonlee@cs.toronto.edu})
    }
}

\input{figures}

\begin{document}
\maketitle
\thispagestyle{empty}
\pagestyle{empty}

\begin{abstract}
    Planning for sequential robotics tasks often requires integrated symbolic and geometric reasoning. \gls{TAMP} algorithms typically solve these problems by performing a tree search over high-level task sequences while checking for kinematic and dynamic feasibility. This can be inefficient because, typically, candidate task plans resulting from the tree search ignore geometric information. This often leads to motion planning failures that require expensive backtracking steps to find alternative task plans. We propose a novel approach to \gls{TAMP} called Stein Task and Motion Planning (STAMP) that relaxes the hybrid optimization problem into a continuous domain. This allows us to leverage gradients from differentiable physics simulation to fully optimize discrete and continuous plan parameters for \gls{TAMP}. In particular, we solve the optimization problem using a gradient-based variational inference algorithm called Stein Variational Gradient Descent. This allows us to find a distribution of solutions within a single optimization run. Furthermore, we use an off-the-shelf differentiable physics simulator that is parallelized on the GPU to run parallelized inference over diverse plan parameters. We demonstrate our method on a variety of problems and show that it can find multiple diverse plans in a single optimization run while also being significantly faster than existing approaches. \href{https://rvl.cs.toronto.edu/stamp}{\texttt{https://rvl.cs.toronto.edu/stamp}}
\end{abstract}

\section{INTRODUCTION}
Task and Motion Planning (\gls{TAMP}) is central to many sequential decision-making problems in robotics, which often require integrated logical and geometric reasoning to generate a feasible symbolic action and motion plan that achieves a particular goal \cite{garrett2021integrated}. In this paper, we present a novel algorithm called \gls{STAMP}, which uses \gls{SVGD} \cite{Liu2016-ee, han2020stein}, a variational inference method, to efficiently generate a distribution of optimal solutions in a single run. Unlike existing methods, we transform \gls{TAMP} problems, which operate over both discrete symbolic variables and continuous motion variables, into the continuous domain. This allows us to run gradient-based inference using \gls{SVGD} and differentiable physics simulation to generate a diversity of plans.

Prior works such as \cite{Garrett2020-ts, Toussaint2015-ye, Ortiz-Haro_Karpas_Toussaint_Katz_2022, ren2021extended, Ren_Chalvatzaki_Peters_2021} solve \gls{TAMP} problems by performing a tree search over discrete logical plans and integrating this with motion optimization and feasibility checking. By leveraging gradient information, \gls{STAMP} forgoes the need to conduct a computationally expensive tree search that might involve backtracking and might be hard to parallelize. Instead, \gls{STAMP} infers the relaxed logical action sequences jointly with continuous motion plans, without a tree search. 

\begin{figure}
    \centering
    \includegraphics[width=\linewidth]{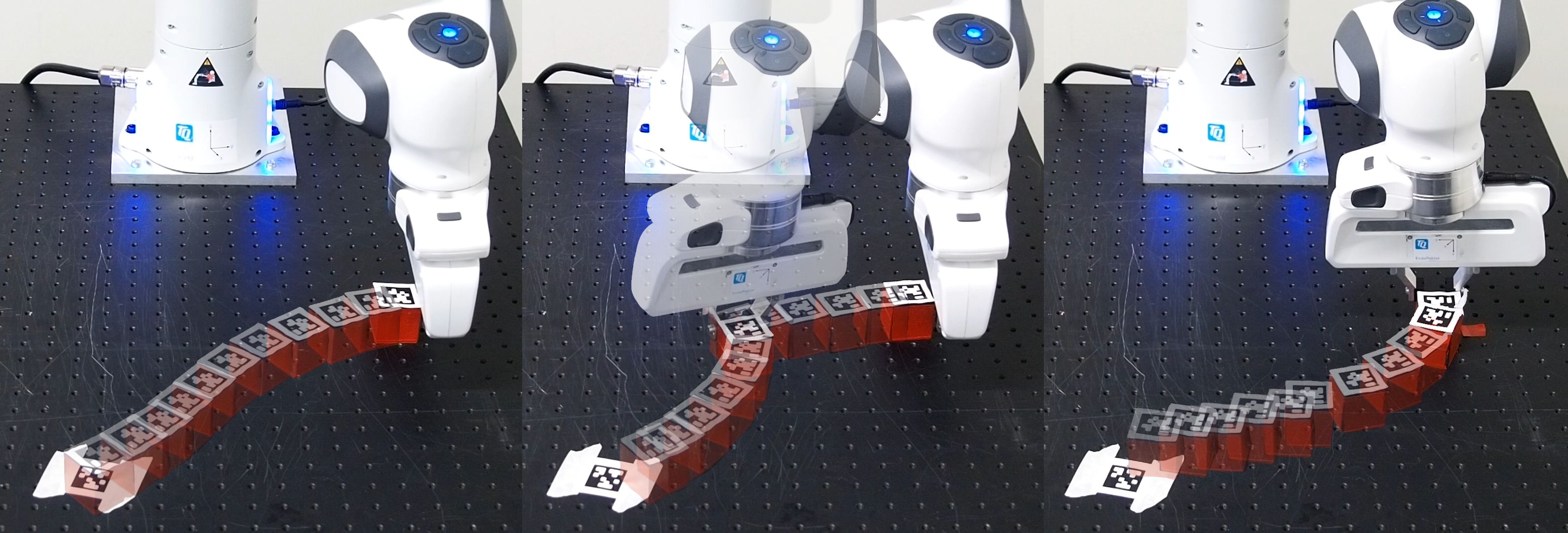}
    \setlength{\abovecaptionskip}{-0.4cm} 
    \caption{Demonstrations of the top-3 block-pushing plans found by STAMP on a Franka manipulator. Task plans from left to right: push the East side; push the East side, then the North side; push the North side. STAMP found all three solutions in one run.}
    \vspace{-0.5cm} 
    \label{fig:real_pusher}
\end{figure}

Further, by solving a Bayesian inference problem over the search space and utilizing GPU parallelization, \gls{STAMP} conducts a parallelized optimization over multiple logical and geometric plans at once. As a result, it produces large, diverse plan sets that are crucial in downstream tasks with replanning, unknown user preferences, or uncertain environments. While several diverse planning methods have been developed for purely symbolic planning \cite{speck-et-al-aaai2020,katz-et-al-icaps2018}, most \gls{TAMP} algorithms do not explicitly solve for multiple plans and suffer from an exploration-versus-computation time trade-off. 

Why would we need multiple solutions if only one will eventually be executed? One reason is that having access to a diverse set of solutions provides added flexibility in selecting a feasible plan based on criteria that were not initially considered. Second, the push for diverse solutions might lead to unexpected, but feasible plans. Third, when TAMP is used as a generating process for training data for imitation learning~\cite{dalal2023optimus}, a diverse set of solutions is preferable to learn policies that induce multi-modal trajectory distributions.     

\figStampPipeline

The main contributions of our work are (a) introducing a relaxation of the discrete symbolic actions and thus reformulating \gls{TAMP} as an inference problem on continuous variables to avoid tree search; (b) solving the resulting problem with gradient-based \gls{SVGD} inference updates using an off-the-shelf differentiable physics simulator; and (c) parallelizing the inference process on the GPU, so that multiple diverse plans can be found in one optimization run. 

\section{Related Work}
\label{sec:related_work}

\textbf{Task and Motion Planning} Guo et al.~\cite{guo2023} categorize \gls{TAMP} solvers into sampling-based methods \cite{Garrett2020-ts, garrett2018ffrob, garrett2018sampling, garrett2018stripstream, garrett2020pddlstream, thomason2019unified}; hierarchical methods \cite{kaelbling2011hierarchical, Kaelbling2013-yl, suarez2018interleaving}; constraint-based methods \cite{dantam2016incremental, dantam2018incremental, he2015towards}; and optimization-based methods \cite{Toussaint2015-ye, zhang2016co, saha2017task}. Our method falls under optimization-based methods, which find a complete task and motion plan that optimizes a predefined cost function

\textbf{Probabilistic Planning} Our work is inspired by the idea of approaching planning as inference \cite{botvinick2012planning}. Prior works such as \cite{shah2020anytime, kaelbling2013integrated, Ha2020-mu, csucan2012accounting, zhao2021hierarchical, kaelbling2011domain, hadfield2015modular, hou2021belief, garrett2020online} have explored the intersection of probability and \gls{TAMP}. 
Ha et al.~\cite{Ha2020-mu} developed probabilistic Logic Geometric Programming (LGP) for solving \gls{TAMP} in stochastic environments, Shah et al.~\cite{shah2020anytime} developed an anytime algorithm algorithm for \gls{TAMP} in stochastic environments, and Kaelbling and Lozano-P\'erez extended Hierarchical Planning in the Now \cite{kaelbling2011hierarchical} to handle current and future state uncertainty. \gls{STAMP}'s probabilistic interpretation is similar to \cite{Ha2020-mu,Lambert2020-tg} in that we run inference over a posterior plan distribution, but differs in that our distribution is defined over both discrete and continuous plan parameters rather than over only continuous parameters. While many stochastic \gls{TAMP} methods can be computationally expensive \cite{guo2023}, our method runs efficient, gradient-based inference through parallelization.

\textbf{Diverse Planning} Diverse or top-k symbolic planners like \textsc{sym-k} and \textsc{forbid-k} are used to produce sets of feasible task plans \cite{speck-et-al-aaai2020, katz-et-al-icaps2018}. Existing work in \gls{TAMP} generate different logical plans by adapting diverse symbolic planners, which are iteratively updated with feasibility feedback from the motion planner\cite{Ortiz-Haro_Karpas_Toussaint_Katz_2022, Ren_Chalvatzaki_Peters_2021}. Ren et al.~\cite{Ren_Chalvatzaki_Peters_2021} rely on a top-k planner to generate a set of candidate logical plans, but only to efficiently find a single \gls{TAMP} solution rather than a distribution of diverse plans. More similarly to \gls{STAMP}, Ortiz et al.~\cite{Ortiz-Haro_Karpas_Toussaint_Katz_2022} seek to generate a set of plans based on a novelty criteria, but enforce task diversity by iteratively forbidding paths in the logical planner. In contrast, \gls{STAMP} finds diverse plans by solving \gls{TAMP} as an inference problem over plan parameters.

\textbf{Differentiable Physics Simulation \& TAMP} Differentiable physics simulators \cite{De_Avila_Belbute-Peres2018-sm, Qiao2020-gu, Werling2021-ui, Howell2022DojoAD, Jatavallabhula2021-go, warp2022, brax, hu2019difftaichi} solve a mathematical model of a physical system while allowing the computation of the first-order gradient of the output directly with respect to the parameters or inputs of the system. 
They have been used to optimize trajectories \cite{Howell2022DojoAD}, controls \cite{Heiden2021DiSECtAD}, or policies \cite{Xu2022AcceleratedPL}; and for system identification \cite{Heiden2021-sl, Jatavallabhula2021-go}. Toussaint et al.~\cite{Toussaint2018-em} have used differentiable simulation within LGP for sequential manipulation tasks by leveraging simulation gradients for optimization at the path-level. In contrast, \gls{STAMP} uses simulation gradients to optimize both symbolic and geometric parameters.

Envall et al.~\cite{envalldifferentiable} used gradient-based optimization for task assignment and motion planning. Their problem formulation allows task assignments to emerge implicitly in the solution. In contrast, we optimize the task plan explicitly through continuous relaxations, use gradient-based inference to solve \gls{TAMP}, and obtain gradients from differentiable simulation.

\section{Preliminaries}
\label{sec:background}

\subsection{Stein Variational Gradient Descent}
\Gls{SVGD} is a variational inference algorithm that uses particles to fit a target distribution. Particles are sampled randomly at initialization and updated iteratively until convergence, using gradients of the target distribution with respect to each particle \cite{Liu2016-ee}. \Gls{SVGD} is fast, parallelizable, and able to fit both continuous \cite{Liu2016-ee} and discrete \cite{han2020stein} distributions.

\subsubsection{SVGD for Continuous Distributions \cite{Liu2016-ee}}
Given a target distribution $p(\theta)$, $\theta\in\mathbb{R}^d$, a randomly initialized set of particles $\{\theta_i\}_{i=1}^n$, a positive definite kernel $k(\theta,\theta')$, and step size $\epsilon$, \gls{SVGD} iteratively applies the following update rule on $\{\theta_i\}_{i=1}^n$ to approximate the target distribution $p(\theta)$ (where $k_{ji} = k(\theta_j, \theta_i)$ for brevity):
\begin{equation}
    \theta_i \leftarrow \theta_i + \frac{\epsilon}{n}\sum_{j=1}^n \bigl[ 
   \underbrace{\nabla_{\theta_j} \log p(\theta_j)k_{ji}}_{\text{(A)}}  + \underbrace{\nabla_{\theta_j}k_{ji}}_{\text{(B)}} \bigr]
    \label{eq:svgd}
\end{equation}
Term (A), which is a kernel-weighted gradient, encourages the particles to converge towards high-density regions in the target distribution. Term (B) induces a ``repulsive force" that prevents all particles from collapsing to a maximum \textit{a posteriori} solution, i.e., it encourages exploration while searching over continuous parameters. This property allows \gls{STAMP} to find multiple diverse solutions in parallel.

\subsubsection{SVGD for Discrete Distributions \cite{han2020stein}}
Given a target distribution $p_*(z),z\in\mathcal{Z}$ on a discrete set $\mathcal{Z}$, \gls{DSVGD} introduces relaxations to $\mathcal{Z}$ that reformulates discrete inference as inference on a continuous domain. In our case, $z$ denotes a symbolic/discrete action. To do the relaxation, we construct a differentiable and continuous surrogate distribution $\rho(\theta), \theta\in\mathbb{R}^d$, that approximates the discrete distribution $p_*(z)$. Crucially, a map $\Gamma: \mathbb{R}^d\rightarrow\mathcal{Z}$ is defined such that it divides an arbitrary base distribution $p_0(\theta), \theta\in\mathbb{R}^d$ (e.g., a Gaussian or uniform distribution) into $K$ partitions with equal probability. That is, 
 \begin{equation}
    \int_{\mathbb{R}^d} p_0(\theta) \, \mathbb{I}[z_i=\Gamma(\theta)] \,\mathrm{d}\theta = 1/K\label{eq:evenpartition}
\end{equation}
Then, the surrogate distribution $\rho(\theta)$ can be defined as \(\rho(\theta) \propto p_0(\theta) \tilde{p}_*\big(\tilde{\Gamma}(\theta)\big)\)~\cite{han2020stein}, where $\Tilde{p}_*(\theta),$ $\theta\in\mathbb{R}^d$ simply denotes ${p}_*(z),$ $z\in\mathcal{Z}$ defined on the continuous domain. $\Tilde{\Gamma}(\theta)$ is a smooth relaxation of ${\Gamma}(\theta)$; for instance, $\Tilde{\Gamma}(\theta)=\text{softmax}(\theta)$ is a smooth relaxation of ${\Gamma}(\theta)=\text{max}(\theta)$. After initializing particles $\{\theta_i\}_{i=1}^N$ \textit{defined on the continuous domain}, \Gls{DSVGD} uses the surrogate to update $\{\theta_i\}_{i=1}^N$ via $\theta_i\leftarrow\theta_i+\epsilon\Delta\theta_i$, where $k_{ji} = k(\theta_j, \theta_i)$, $w = \sum_i w_i$, $\rho_j=\rho(\theta_j)$, and:
\begin{equation}
        \Delta\theta_i = \sum_{j=1}^{n} \frac{w_j}{w} \big(\nabla_{\theta_j}\log\rho_j + \nabla_{\theta_j}\big)k_{ji} ,\
        w_j = \frac{\Tilde{p}_*\big(\Tilde{\Gamma}(\theta_j)\big)}{p_*\big(\Gamma(\theta_j)\big)}  
        \label{eq:dsvgd}
\end{equation}
Post-convergence, the discrete counterpart of each particle can be recovered by evaluating $\{z_i = \Gamma(\theta_i)\}_{i=1}^N$.

\subsection{Differentiable Physics Simulation}
Physics simulators solve the following dynamics equation
\begin{equation}
    M\ddot{x} = J^\top f(x,\dot{x}) + C(x,\dot{x}) + \tau(x,\dot{x},a)
\end{equation}
to roll out the state (position and orientation) $x$ of the system, subjected to external forces $f$, Coriolis force $C$, and joint actuations $\tau$. The Warp simulator \cite{warp2022} used in our paper supports time integration of the above via Semi-Implicit Euler \cite{erez2015simulation} and XPBD \cite{macklin2016xpbd, macklin2019small} schemes and allows fast forward simulations and gradient computations through GPU parallelization. We use the semi-implicit Euler integrator due to its numerical stability \cite{erez2015simulation, Jatavallabhula2021-go} and demonstrated success in similar problems \cite{Xu2022AcceleratedPL, macklin2019small}. Contacts are resolved in Warp using a spring-based non-penetrative model proposed in \cite{xu2021end, Xu2022AcceleratedPL}, while joint limits for articulated bodies are enforced with a spring model following the approach in \cite{Xu2022AcceleratedPL}.

Warp computes a forward pass to roll out the system's states $x_{t+1} = f_\text{sim}(x_{t},\theta_t)$ for all time steps $t=1,...,T$ given control and plan parameters $\theta_t$. Given a loss $\mathcal{L}$ defined on the final or intermediate states, gradients can be backpropagated through the entire trajectory with respect to simulation and plan parameters. As we will explain in following sections, we leverage the differentiability of the simulator to compute the gradients $\nabla_{\theta_j} \log p(\theta_j)$ in term (A) of equation \eqref{eq:svgd}.

\figMixtureOfGaussians
\figBilliardSolutions

\section{Our Method: Stein Task and Motion Planning}
\label{sec:method}
\gls{STAMP} solves \gls{TAMP} as a variational inference problem over discrete symbolic actions and continuous motion plans, conditioned on them being optimal. An analogy to this is trying to sample particles from a known mixture of Gaussians, in which each Gaussian has a distinct class. \gls{SVGD} must move the particles towards high-density regions in the Gaussian (thus learning the continuous parameters) and also learn the correct classes. In \gls{TAMP}, the Gaussian mixture is analogous to a target distribution where higher likelihood is assigned to optimal solutions, the discrete class is analogous to the discrete task plan, and the continuous parameters are analogous to the motion plan. Figure \ref{fig:svgd_gm} illustrates how, after \gls{SVGD}, particles with the same color usually belong to the same Gaussian.

\subsection{Problem Formulation}  
\label{sec:problem-formulation}

Given a problem domain, we are interested in sampling optimal symbolic/discrete actions $z_{1:K}\in\mathcal{Z}^K$ and continuous controls $u_{0:KT-1}\in\mathbb{R}^{KT}$ from the posterior
\begin{equation}    
    \label{eq:og_posterior}
    p(z_{1:K}, u_{0:KT-1} \mid O = 1) 
\end{equation}
where $O \in \{0, 1\}$ indicates optimality of the plan; $K$ is the number of symbolic action sequences; $T$ is the number of timesteps by which we discretize each action sequence; and $\mathcal{Z}$ is a discrete set of all $m$ possible symbolic actions that can be executed in the domain (i.e., $|\mathcal{Z}|=m$). $z_{1:K}$ and $u_{0:KT-1}$ fully parameterize a task and motion plan, since they can be inputted to a physics simulator $f_\text{sim}$ along with the initial state $x_0$ to roll out the system's state at every timestep, that is, $x_{1:KT}=f_\text{sim}(x_0, z_{1:K},u_{0:KT-1})$\footnote{We later introduce $a_{1:K}$, a continuous relaxation of $z_{1:K}$. The forward simulation can also be rolled out using $a_{1:K}$ as $x_{1:KT}=f_\text{sim}(x_0, a_{1:K},u_{0:KT-1})$, which will ensure gradients can be taken with respect to $a_{1:K}$. This is important as $z_{1:K}$ is discrete.}.

We use \gls{SVGD} to sample optimal plans from $p(z_{1:K}, u_{0:KT-1} \mid O = 1)$. As $p$ is defined over both continuous and discrete domains, to use \gls{SVGD}'s gradient-based update rule, we follow Han et al.~\cite{han2020stein}'s approach and construct a differentiable surrogate distribution for $p$ defined over a continuous domain. We denote this surrogate by $\rho$. 

To construct the surrogate, we first relax the discrete variables by introducing a map from the real domain to $\mathcal{Z}^K$ that evenly partitions some base distribution $p_0$. For any problem domain with $m$ symbolic actions, we define $\mathcal{Z}=\{e_i\}_{i=1}^m$ as a collection of $m$-dimensional one-hot vectors\footnote{The $i$th element of $e_i$ is 1 and all other elements are 0.}. Then, $z_{1:K}\in\mathcal{Z}^K$, implying that we commit to one action ($e_i$) in each of the $K$ action phases. Further, we use a uniform distribution for the base distribution $p_0$. Given this formulation, we can define our map\footnotemark $\;\Gamma:\mathbb{R}^{mK}\rightarrow\mathcal{Z}^K$ and its differentiable surrogate $\Tilde{\Gamma}$ as the following:
\begin{align}
    \label{eq:gamma_map}
    \Gamma(a_{1:K}) &= [\max\bigl\{a_1\bigr\} \ldots \max\bigl\{a_K\bigr\}]^\top  = z_{1:K}\\ \nonumber
    \Tilde{\Gamma}(a_{1:K}) &= [\text{softmax}\bigl\{a_1\bigr\} \ldots \text{softmax}\bigl\{a_K\bigr\}]^\top = \tilde{z}_{1:K}.   
\end{align}
We show in Appendix \ref{app:dsvgd_derivations} that the above map indeed partitions $p_0$ evenly when $p_0$ is the uniform distribution.
\footnotetext{Given $a_k\in\mathbb{R}^m$, $\max\bigl\{a_k\bigr\}$ returns a $m$-dimensional one-hot encoding $e_i$ iff the $i$th element of $a_k$ is the larger than all other elements.} By constructing the above map, we can run inference over purely continuous variables $a_{1:K}\in\mathbb{R}^{mK}$ and $u_{0:KT-1}\in\mathbb{R}^{KT-1}$ and recover the discrete plan parameters post-inference via $z_{1:K} = \Gamma(a_{1:K})$. We denote $a_{1:K}\in\mathbb{R}^{mK}$ and $u_{0:KT-1}\in\mathbb{R}^{KT-1}$ collectively as a particle, that is, $\theta=[a_{1:K}, u_{0:KT-1}]^\top$, and randomly initialize $\{\theta_i\}_{i=1}^n$ to run SVGD inference. The target distribution we aim to infer is a differentiable surrogate $\rho$ of the posterior $p$ defined as:
\begin{gather}
    \label{eq:surrogate_def}
    \rho(a_{1:K},u_{0:KT-1} \mid O=1)  \\
    \propto p_0(a_{1:K}) \, p(\widetilde{\Gamma}(a_{1:K}),u_{0:KT-1} \mid O=1).
\end{gather}

\noindent In practice, because $p_0(a_{1:K})$ can be treated as a constant by making its boundary  arbitrarily large (see Appendix \ref{subapp:base_dist}), we simply remove $p_0$ from the expression above.

Note that it is not necessary to normalize the $\rho$ as the normalization constant vanishes by taking the gradient of $\log\rho$ in the \gls{SVGD} updates. We now quantify the surrogate of the posterior distribution. We begin by applying Bayes' rule and obtain the following factorization of $\rho(\theta \mid O = 1)$:
\begin{gather}
    \rho(\theta \mid O = 1) 
    \propto p(O = 1 \mid \theta) \, p(\theta)
    \label{eq:posterior_dist}
\end{gather}
Since the likelihood function is synonymous with a plan's optimality, we formulate the likelihood via the total cost $C$ associated with the relaxed task and motion plan $\theta$:
\begin{equation}
    \begin{gathered}
        p(O = 1 \vert \theta)  
        \propto \exp{-C(\theta, x_{0:KT})}
    \end{gathered}
    \label{eq:neglog_likelihood}
\end{equation}

\noindent Meanwhile, the prior $p(\theta)$ is defined over $\theta$ only and as such, we use it to impose constraints on the plan parameters (e.g., kinematic constraints on robot joints). 

Related to the \gls{LGP} \cite{Toussaint2018-em, Driess2022-tp, Toussaint2020, Toussaint2015-ye, Ha2020-mu} framework, which optimizes a cost subject to constraint functions that activate or deactivate when kinematic/logical action transition events occur, $C(\theta)$ in our method is a sum of \textit{relaxed} versions of both the cost and constraint functions. Here, ``relaxed'' refers to removing the cost and contraints' dependence on the discrete logical variables through our relaxations, and defining costs/constraint functions that are differentiable w.r.t. our plan parameters $\theta$.

\subsection{Problem Reduction Using Motion Primitives}
\label{sec:dmp_for_stamp}
Rather than running inference over $\theta = [a_{1:K}, u_{0:KT-1}]^\top$, we can also reduce the dimensionality of $\theta$ using a bank of goal-conditioned \gls{DMP}s (more details in Appendix \ref{app:dmp}). We redefine the particle to be $\theta=[a_{1:K}, g_{1:K}]^\top$, where $g_{1:K}\in\mathbb{R}^{Kd}$ denotes goal poses for the system after executing action $k$, and $d$ is the DOF of the system. With just $g_k$, we recover the full trajectory $x_{0:KT-1}$ by integrating the system of equations in \eqref{eq:canonicalSystem} and computing the controls $u_{0:KT-1}$ using a simple trajectory tracking controller. Note that to do this, we train a bank of goal-conditioned \gls{DMP}s from demonstration data for every action offline. This new inference problem over $[a_{1:K}, g_{1:K}]^\top$ is more tractable than the original inference problem over $[a_{1:K}, u_{0:KT-1}]^\top$ due to the reduced dimensionality.

\begin{figure}[t!]
    \centering
    \vspace{0.2cm}
    \includegraphics[width=0.95\linewidth, trim={0.3cm 0.4cm 0 0}, clip]{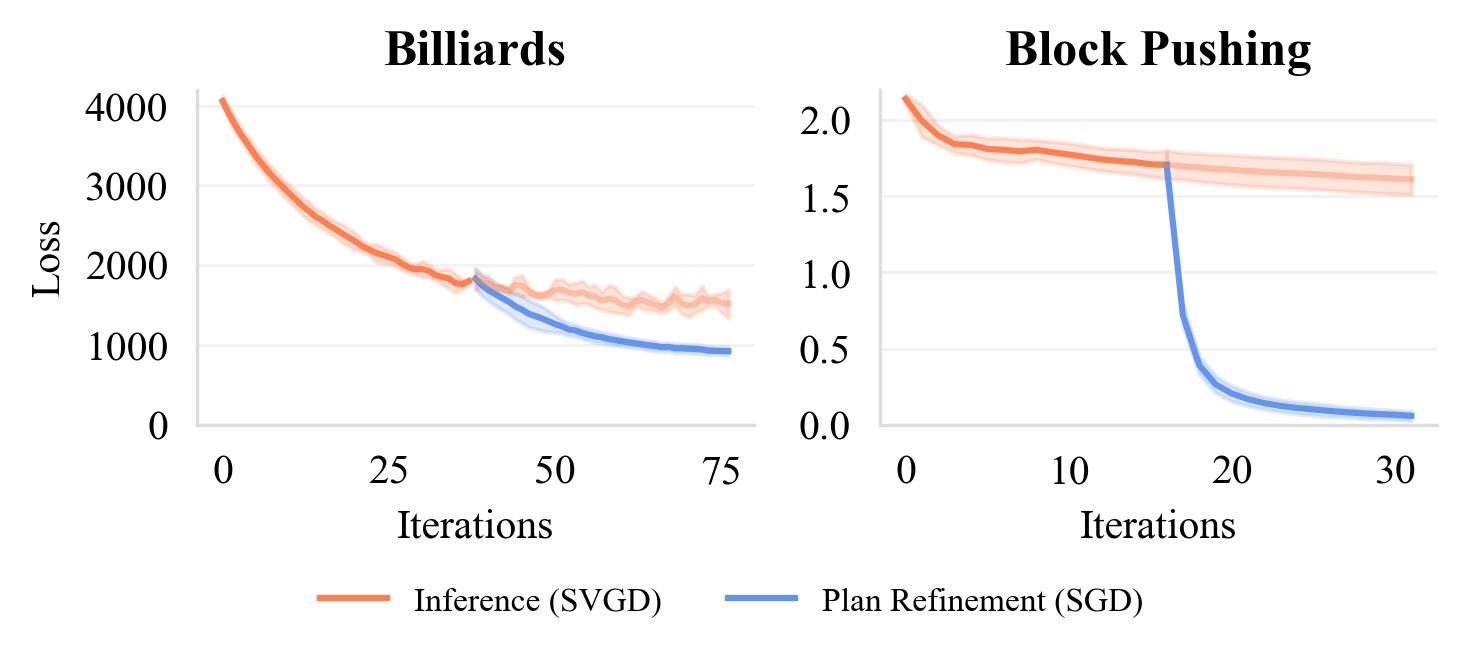}
    \setlength{\abovecaptionskip}{-0.cm} 
    \caption{Evolution of mean cost over all particles averaged across 5 runs.}
    \vspace{-0.5cm} 
    \label{fig:loss_evolution}
\end{figure}

\subsection{STAMP Algorithm}
\label{sec:stamp_algo_description}
Given a problem domain with cost $C(\theta)$ and $n$ randomly initialized particles $\theta=\{\theta_i\}_{i=1}^n$, \gls{STAMP} finds a distribution of optimal solutions to the inference problem $\rho(\theta | O=1)$ by running \gls{SVGD} inference until convergence and subsequently refining the plans via SGD updates. During \gls{SVGD}, each of the $n$ plans ($\theta$) are executed and simulated in Warp. After obtaining states $x_{0:KT} = f_\text{sim}(x_0, a_{1:K},u_{0:KT-1})$ for each particle, the posterior distribution and its surrogate are computed using equations \eqref{eq:og_posterior} and \eqref{eq:surrogate_def}. Gradients of the log-posterior with respect to $\{\theta_i\}_{i=1}^n$ are obtained via Warp's auto-differentiation capabilities, and these gradients are used to update each candidate plan $\theta_i$ via update rule \eqref{eq:dsvgd}. During \gls{SGD}, the above steps are repeated except for the update rule, which changes to $\theta \leftarrow \theta + \epsilon\nabla_\theta\log\rho$. Switching to \gls{SGD} after \gls{SVGD} allows us to finetune our plans, as the repulsive term $\nabla_\theta k(\theta,\theta')$ in \gls{SVGD} can push the plans away from optima, but with \gls{SGD} the plans are optimized to reach them. All of the above computations (physics simulations, log posterior evaluations, and gradient computations for all particles) are run in parallel on the GPU, making \gls{STAMP} highly efficient. Figure \ref{fig:pipeline} shows the algorithm pipeline and Algorithm \ref{algo:svgd-sgd} shows the pseudocode.

\figHists

\section{Evaluation}
\label{sec:results}

We run \gls{STAMP} on three problems: billiards, block-pushing, and pick-and-place. We benchmark \gls{STAMP} against two \gls{TAMP} baselines that build upon PDDLStream \cite{Garrett2020-ts} and Diverse \gls{LGP} \cite{Ortiz-Haro_Karpas_Toussaint_Katz_2022} (implementation details are in Appendix \ref{app:baselines}).


\subsection{Problem Environments and Their Algorithmic Setup}
\label{subsection:problem_envs}

\subsubsection{Billiards}

The goal is to optimize the initial velocity $u_0$ on the cue ball that sends the target ball into one of the two pockets in Figure \ref{fig:billiards_setup}. This requires planning on the continuous domain ($u_0\in\mathbb{R}^2$) and discrete domain (the set of walls we wish for the cue ball to hit before colliding with the target ball). We define our particles as $\theta=[u_0,z]$ where $z = [z_1, z_2, z_3, z_4] \in \{ 0, 1 \}^4$, which indicates which of the four walls the cue ball bounces off of.\footnote{$z_i=1$ if the cue ball hits wall $i$, and it is $0$ otherwise.} The cost function $C$ is a weighted sum of the target loss $L_\text{target}$ and aim loss $L_\text{aim}$, which are minimal when the cue ball ends up in either of the two pockets if the cue ball comes in contact with the target ball at any point in its trajectory, respectively. Hyperparameters $\beta_\text{target},\beta_\text{aim}>0$ are the loss' respective coefficients:
\begin{equation}
	C(\theta) = \beta_\text{aim}L_\text{aim} + \beta_\text{target}L_\text{target} 
\end{equation}
Appendix \ref{app:billiards_setup} provides additional details on the problem setup.

\subsubsection{Block-Pushing}

The goal is to push the block in Figure \ref{fig:blockpusher_setup} towards the goal region. The symbolic plan is the sequence of sides (north, east, south, west) to push against. Assuming \textit{a priori} that up to $K$ action sequences can be committed, we define the particles as $\theta = \bigl[ a_1, \dots, a_K, g_1, \dots, g_K \bigr]^\top$, where $a_k\in\mathbb{R}^4$ are the relaxed symbolic variables and $g_k=[g_k^x,g_k^y,g_k^\phi]\in\mathbb{R}^{3}$ denote individual goal poses after executing each action. Given the goals, we use pretrained \gls{DMP}s to obtain trajectories and control inputs. The discrete task variable for the $k\/$th action can be recovered via equation \eqref{eq:gamma_map}. 
The cost is the weighted sum of the target loss $L_\text{target}$ and trajectory loss $L_\text{traj}$. $L_\text{target}$ penalizes large distances between the cube and the goal region at the final time-step. 
\begin{equation}
    C(\theta) = \beta_\text{target}L_\text{target}(x_{KT}) + \beta_\text{traj}L_\text{traj}(g_{1:K}^{x,y})
\end{equation}
Appendix \ref{app:pusher_setup} provides further details on the problem setup.

\begin{table*}
    \addtocounter{table}{-1}
    \centering
    \subfloat{\tabRuntimeComparison}
    \subfloat{\tabStampRuntimeParticles}
    \subfloat{\tabStampRuntimeDimensionality}
    \vspace{-0.2cm} 
\end{table*}

\subsubsection{Pick-and-Place}
\label{subsection:method-pick-place}

\begin{figure*}[!t]
    \centering
    {\includegraphics[width=\linewidth]{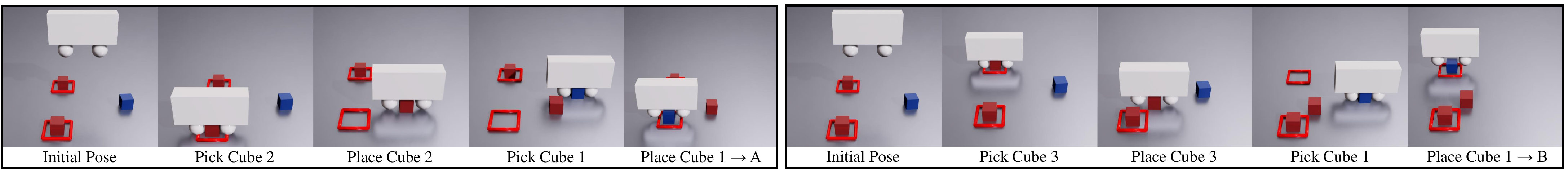}}
    \setlength{\abovecaptionskip}{-0.4cm} 
    \caption{Both solutions for the pick and place experiment obtained simultaneously from running STAMP. See Figure \ref{fig:bs_setup} for the experiment setup. The left solution removes Cube 2 to place Cube 1 in target A. The right solution removes Cube 3 to place Cube 1 in target B.}
    \vspace{-0.5cm} 
    \label{fig:bs_solutions}
\end{figure*}

Using an end effector, the goal is to pick and place blocks 1, 2, and 3 such that block 1 ends up in one of two targets shown in Figure \ref{fig:bs_setup}. Long horizon reasoning is required in this problem, as blocks 2 and 3 occlude the targets; hence, either block 2 or 3 have to be moved out of the way before block 1 can be placed. 
Our goal can be expressed as
\begin{gather}
    \big((\texttt{cube1} \in \mathcal{A}) \lor (\texttt{cube1} \in \mathcal{B})\big) \\ \nonumber
    \land \lnot\texttt{onTop(cube1,cube2)} \land \lnot\texttt{onTop(cube1,cube3)}
\label{eq:pp_goal_symbolic}
\end{gather}
Symbolic actions are $\mathcal{Z}=\{\texttt{pick(cube), \texttt{place(cube)}}\}$ $\forall \texttt{cube}\in\mathcal{C}$ (where $\mathcal{C}=\{\texttt{cube1}, \texttt{cube3}, \texttt{cube3}\}$) which are represented numerically via one-hot encodings as discussed in Section \ref{sec:problem-formulation}. Particles are defined as $\theta = \bigl[ a_1, \dots, a_K, g_1, \dots, g_K \bigr]^\top$, where $a_{1:K}$ represent which $K$-length sequence of symbolic actions are executed, and $g_{1:K}$ represent the intermediate goal poses of the end effector after executing each action.

Unlike in previous examples, the symbolic actions have preconditions and postconditions which constitute constraints in our problem. For each pre/postcondition, we construct differentiable loss functions $L_{\texttt{pre(}\cdot\texttt{)}}, L_{\texttt{post(}\cdot\texttt{)}} \geq 0$ that are minimal when the condition is met. The constraint associated with action $z \in \mathcal{Z}$ is then expressed as the $\gamma_i$- and $\gamma_j$-weighted sum of its precondition losses and postcondition losses, where $\gamma_i,\gamma_j>0$ are scalar hyperparameters:
\begin{equation}
    L_z(x_{t_0:t_T}) = \sum_{i\in\texttt{pre(}z\texttt{)}} \gamma_i L_i (x_{t_0},g)
    + \sum_{j\in\texttt{post(}z\texttt{)}} \gamma_j L_j (x_{t_T})
    \label{eq:pre_post_constraints}
\end{equation}
Please see Appendix \ref{subsection:prepost_defn}
 for more details on the pre/postcondition losses for each action.

We optimize towards the logical goal stated in \eqref{eq:pp_goal_symbolic} by defining the total target loss as the sum of each cube's individual target loss weighted by a soft indicator function $\omega_c$ that gets activated when the cube is held (see Appendix \ref{app:holding_like_def}). The target loss of cube $c$, $L_{\text{target,}c}$, and its gradient will dominate the optimization if $c$ is held by the gripper, which is a necessary condition for placing $c$ in its target.
\begin{equation}
    L_\text{target}(x_{t_0}, x_{t_T}) = \sum_{c\in\mathcal{C}} \omega_c(x_{t_0}) \cdot L_{\text{target,}c}(x_{t_T}).
\end{equation}
The target loss for each cube is defined in Appendix \ref{app:pick_place_target_loss}.

Finally, the total cost function is constructed as a sum over all constraints $\mathbf{L}_z(x_{t_0:t_T})=[L_z(x_{t_0:t_T}) \forall z\in\mathcal{Z}]^\top$ weighted by $\tilde{z_k} = \tilde{\Gamma}(a_k) \forall k=1,...,K$ and the target loss, which is computed after executing each action\footnote{Recall that $K$ is the number of action sequences, $a_k$ the component of the particle representing the task plan, and $\tilde{\Gamma}(\cdot)$ a composition of softmax operations.}.
\begin{equation}
    \label{eq: cost_bs}
    C(\theta) = \sum_{k=1}^K \Big(\tilde{z}_k \cdot \mathbf{L}_z (x_{(k-1)T+1:kT}) + L_{\text{target}}(x_{kT})\Big)
\end{equation}
The global optima of $C(\theta)$ are thus action sequences $z_{1:K}=\Gamma(a_{1:K})$ and intermediate goal poses that solve the goal while ensuring all pre/postconditions are met. Further, $C$ is amenable to gradient-based methods like STAMP since it is differentiable w.r.t. both $a_{1:K}$ and $g_{1:K}$.
Note that because of the construction of $C(\theta)$ as a sum over $k$, STAMP can be run for each $k$ separately (see Appendix \ref{subsection:block_opt}), diminishing the need to take gradients through the entire trajectory $x_{0:KT}$, which may be prone to exploding gradients.


\subsection{Experimental Results (Simulation)}

We investigate the following questions: 
\begin{enumerate}
    \item[(Q1)] Can \gls{STAMP} return a variety of plans to problems for which multiple solutions are possible?  
    \item[(Q2)] How does \gls{STAMP}'s runtime compare to search-based \gls{TAMP} baselines?
    \item[(Q3)] How does \gls{STAMP}'s runtime scale w.r.t. problem dimensionality and number of particles?
    \item[(Q4)] Is SGD necessary in \gls{STAMP} to find optimal plans?
\end{enumerate}

\subsubsection{Solution Diversity}
\gls{STAMP} finds a variety of solutions to all three problems; Figures \ref{fig:billiards_sols}, \ref{fig:pushing_sols}, and \ref{fig:bs_solutions} show sample solutions. While \gls{STAMP} finds a distribution of solutions in a single run, baseline methods can only find one solution per run. In Figures \ref{fig:histo_billiards} and \ref{fig:histo_pusher}, we show a histogram of solutions \gls{STAMP} found in one run compared to solutions found using baseline methods by invoking them multiple times.

\subsubsection{Runtime Efficiency}
\gls{STAMP} produces large plan sets in substantially less time than the baselines. Table \ref{tab:runtime_baselines} shows that the time it takes \gls{STAMP} to generate hundreds of solutions is often less than the time needed for Diverse LGP \cite{Ortiz-Haro_Karpas_Toussaint_Katz_2022} and PDDLStream \cite{Garrett2020-ts} baselines to produce a single solution.

\subsubsection{Scalability to Higher Dimensions and Greater Number of Particles}
Tables \ref{tab:runtime_num_particles} and \ref{tab:runtime_dimensions} show the total runtime of our algorithm while varying the number of particles and particle dimensions. Increasing the number of particles and the dimensions of the particles has little effect on the total optimization time, which is consistent with expectations as our method is parallelized over the GPU.\footnote{GPU parallelization is made possible through the use of SVGD, a parallelizable inference algorithm, and the Warp simulator.} 

\subsubsection{SGD Plan Refinement}
Figure \ref{fig:loss_evolution} shows the cost evolution with SGD plan refinement versus solely using SVGD for the same number of iterations. Pure SVGD inference results in slow or poor convergence, while running SGD post-SVGD inference results in better convergence. 

\subsection{Experimental Results (Real Robot)}
We demonstrate \gls{STAMP} on a real robot system using a front-view camera, AprilTags \cite{Olson_2011} for pose estimation of cube(s) and target(s), and a Franka Emika Panda robotic arm. We input the detected poses into \gls{STAMP} and track the resulting trajectory directly. For the block pushing experiment, we vary the cube and target positions across 3 distinct configurations and set the end-effector to follow the pose of one of the block's faces. While several trajectories returned by STAMP involve rotations that are infeasible due to differences in real world friction, the lowest loss solution in the most popular 3-5 modes of each problem setup successfully solve the problem, as shown in Figure \ref{fig:real_pusher}. More real world experiments and videos can be found on our \href{https://rvl.cs.toronto.edu/stamp}{\textbf{project website}}.

\section{Conclusion}
We introduced STAMP, a novel algorithm that approaches \gls{TAMP} as a variational inference problem over discrete symbolic action and continuous motion parameters. \gls{STAMP} solves the inference problem using \gls{SVGD} and gradients from differentiable simulation. We validated our approach on the billiards, block-pushing, and pick-and-place problems, where \gls{STAMP} was able to discover a diverse set of plans covering multiple different task sequences and motion plans. \gls{STAMP}, through exploiting parallel gradient computation from a differentiable simulator, is much faster at finding a variety of solutions than baselines, and its runtime scales well to higher dimensions and more particles. 

\section{Acknowledgements}
YL was supported by the Google DeepMind Fellowship, NSERC Canada Graduate Scholarship, and Ontario Graduate Scholarship. This article solely reflects the opinions and conclusions of its authors and not the sponsors.

\bibliographystyle{IEEEtran}
\bibliography{IEEEabrv,main}

\newpage
\section*{APPENDIX}
\section{Relaxations for Discrete SVGD}
\label{app:dsvgd_derivations}

Given a problem domain with $m$ possible actions, the goal of \gls{TAMP} is to find a $K$-length sequence of symbolic actions $z_{1:K}\in{Z}^K$ and associated motions that solve some goal. Here, $\mathcal{Z}$ is a discrete set of $m$ possible symbolic actions in the domain. We propose the general formulation of $\mathcal{Z}$ as the set of $m$-dimensional one-hot encodings such that $|\mathcal{Z}|=m$ and $|\mathcal{Z}^K|=m^K$. Then, as per equation \eqref{eq:evenpartition}, we can simply relax $z_{1:K}\in{Z}^K$ to the real domain $\mathbb{R}^{mK}$ by constructing the map $\Gamma:\mathbb{R}^{mK}\rightarrow\mathcal{Z}^K$ as 
\begin{equation}
    \Gamma(a_{1:K}) = \begin{bmatrix}
    \text{max}\bigl\{a_1\bigr\} \\
    \text{max}\bigl\{a_2\bigr\} \\
    \vdots \\
    \text{max}\bigl\{a_K\bigr\}
    \end{bmatrix}
\end{equation}
and its differentiable surrogate as 
\begin{equation}
    \tilde{\Gamma}(\cdot) = \begin{bmatrix}
    \text{softmax}\bigl\{a_1\bigr\} \\
    \text{softmax}\bigl\{a_2\bigr\} \\
    \vdots \\
    \text{softmax}\bigl\{a_K\bigr\}
    \end{bmatrix}.
\end{equation}

We will prove that the above mapping evenly partitions $\mathbb{R}^{mK}$ into $m^K$ parts when the base distribution $p_0$ is a uniform distribution over $[-u,u]^{mK}$ for some $u>0$. We will prove this in two steps: firstly by assuming $K=1$ and then proving this for any positive integer $K$.

\subsection{The Base Distribution}
\label{subapp:base_dist}
Our base distribution $p_0(\theta), \theta\in\mathbb{R}^{mK}$ for discrete \gls{SVGD} is a uniform distribution over the domain $[-u,u]^{mK}$. That is,
\begin{equation}
    p_0(\theta) = \text{unif}(\theta) \\
           = \begin{cases}
                \frac{1}{{(2u)}^{mK}} & \text{if } \theta\in[-u,u]^{mK} \\
                0 & \text{otherwise}.
             \end{cases}    
\end{equation}

Note that in practice, we can make $u\in(0,\infty)$ arbitrarily large to avoid taking gradients of $p_0$ at the boundaries of $[-u,u]^K$.

\subsection{Proof of Even Partitioning: Case $K=1$}
\label{subapp:K_eq_1}
To satisfy the condition that $\Gamma(\theta), \theta\in\mathbb{R}^m$ evenly partitions $\text{unif}(\theta)$, for all $\{e_i\}_{i=1}^m\in\mathcal{Z}$, the following must be true.
\begin{equation}
    \int_{\mathbb{R}^m} \text{unif}(\theta) \mathbb{I}[e_i=\Gamma(\theta)] \,\mathrm{d}\theta = \frac{1}{m}
\end{equation}
Below, we show that the left hand side (LHS) of the above equation simplifies to $\frac{1}{m}$, proving that our relaxation evenly partitions $p_0(\cdot)$.
\begin{align}
    \text{LHS} &= \int_{\mathbb{R}^m} \text{unif}(\theta) \mathbb{I}[e_i=\text{max}(\theta)] \,\mathrm{d}\theta \\
    &= \frac{1}{{(2u)}^{m}} \int_{-u}^u \ldots \int_{-u}^u \mathbb{I}[e_i=\text{max}(\theta)] \,\mathrm{d}x_1 \ldots \mathrm{d}x_m \\
    &= \frac{1}{{(2u)}^{m}} \int_{-u}^u \ldots \int_{-u}^u \prod_{j=1, j\neq i}^{m} \mathbb{I}[x_i>x_j] \,\mathrm{d}x_1 \ldots \mathrm{d}x_m \\
    &= \frac{1}{{(2u)}^{m}} \int_{-u}^{u} \Bigg( \prod_{j=1, j\neq i}^{m} \int_{-u}^{x_i} \mathbb{I}[x_i>x_j] \,\mathrm{d}x_j \Bigg) \,\mathrm{d}x_i \\
    &= \frac{1}{{(2u)}^{m}} \int_{-u}^{u} \Bigg( \prod_{j=1, j\neq i}^{m} \int_{-u}^{x_i} \,\mathrm{d}x_j \Bigg) \,\mathrm{d}x_i \\
    &= \frac{1}{{(2u)}^{m}} \int_{-u}^{u} \Bigg(\int_{-u}^{x_i} \,\mathrm{d}y \Bigg)^{m-1} \,\mathrm{d}x_i \\
    &= \frac{1}{{(2u)}^{m}} \int_{-u}^{u} (x_i+u)^{m-1} \,\mathrm{d}x_i \\
    &= \frac{1}{({2u})^{m}} \cdot \frac{(2u)^m}{m} \\
    &= \frac{1}{m}
\end{align} \qed

\subsection{Proof of Even Partitioning: Case $K>1$}
For $\Gamma(\theta)$ to evenly partition $\text{unif}(\theta), \theta\in\mathbb{R}^{mK}$, the following must be true for all $\{z_i\}_{i=1}^{m^K}\in\mathcal{Z}^K$.
\begin{equation}
    \int_{\mathbb{R}^{mK}} \text{unif}(\theta) \mathbb{I}[z_i=\Gamma(\theta)] \,\mathrm{d}\theta = \frac{1}{m^K}
\end{equation}
We use the results from Appendices \ref{subapp:base_dist} and \ref{subapp:K_eq_1} to show that the left hand side of the above equation simplifies to $\frac{1}{m^K}$, proving that our relaxation evenly partitions $p_0(\cdot)$. Below, we use the following notation: $\vb{u} = [u \ldots u] \in\mathbb{R}^m$, $(z_i)_k$ is the $k$th $m$-dimensional one-hot vector within $z_i\in\mathcal{Z}^K$, and $\vb{\theta}_k = \theta_{(k-1)m+1:km}\in\mathbb{R}^m$.
\begin{align}
    \text{LHS} &= \int_{\mathbb{R}^{mK}} \text{unif}(\theta) \prod_{k=1}^{K} \mathbb{I}[(z_i)_k = \text{max}(\vb{\theta}_k)] \,\mathrm{d}\theta \\
    &= \frac{1}{(2u)^{mK}} \prod_{k=1}^{K} \int_{-\vb{u}}^{\vb{u}} \mathbb{I}[(z_i)_k = \text{max}(\vb{\theta}_k)] \,\mathrm{d}\vb{\theta}_k \\
    &= \frac{1}{(2u)^{mK}} \prod_{k=1}^{K} \frac{(2u)^m}{m} \\
    &= \frac{1}{(2u)^{mK}} \cdot \frac{(2u)^{mK}}{m^K} \\
    &= \frac{1}{m^K}
\end{align}\qed


\section{STAMP Problem Formulation and Methodology}

\subsection{STAMP Pseudocode}
\input{algo}

\subsection{Use of Dynamic Motion Primitives for Problem Reduction}
\label{app:dmp}
Dynamic Motion Primitives (DMP) \cite{Pastor2009-rm} model and generate complex movements by combining a stable dynamical system with transformation functions learned from demonstrations. A movement is modelled with the following system of differential equations:
\begin{align}
    \tau \dot{v} &= K(g-x) - Dv - K(g - x_0)s + Kf(s) \nonumber \\
    \tau \dot{x} &= v \qquad\qquad
    f(s) = \frac{\sum_i w_i \psi_i(s) s}{\sum_i \psi_i(s)} \nonumber\\
    \tau \dot{s} &= -\alpha s  \label{eq:canonicalSystem}
\end{align}
where $x$ and $v$ are positions and velocities; $x_0$ and $g$ are the start and goal positions; $\tau$ is a temporal scaling factor; $K$ and $D$ are the spring and damping constant; $s$ is a phase variable that defines a ``canonical system" in equation \eqref{eq:canonicalSystem}; and $\psi_i(s)$ are typically Gaussian basis functions with different centers and widths. Lastly, $f(s(t))$ is a non-linear function which can be learned to generate arbitrary trajectories from demonstrations. Time-discretized trajectories ${x}(t), {v}(t)$ are generated by integrating the system of differential equations in \eqref{eq:canonicalSystem}.
Thus, learning a DMP from demonstrations can be formulated as a linear regression problem given recorded $x(t)$, $v(t)$, and $\dot{v}(t)$.

At runtime, DMPs can be adapted by specifying task-specific parameters $x_0$, $g$, integrating $s(t)$, and computing $f(s)$, which drives the desired behaviour. 
Our work uses DMPs to generate motion primitives at varying speeds, initial positions, and goals from a small set of demonstrations.   
DMPs can also be extended to incorporate potential fields \cite{Park2008MovementRA, Tan2011APF, Ginesi2020DynamicMP} to avoid collisions, a key requirement for many motion planning tasks.

\section{Billiards Problem Setup}
\label{app:billiards_setup}

\subsection{Graphical Overview}
A graphical depiction of the billiards problem environment is shown in Figure \ref{fig:billiards_setup}.
We recreate this environment in the Warp simulator.

\subsection{Task Variables as Functions of Velocity}
\label{subapp:billiard_wallhits_fn}

We use STAMP to optimize the initial cue ball velocity $u_0$ and the task plan $z = [z_1, z_2, z_3, z_4] \in \{ 0, 1 \}^4$ where $\forall i\in\{1,2,3,4\}$, $z_i=1$ if the cue ball bounces off of wall $i$ and $z_i=0$ otherwise. The walls are labelled 1-4 in Figure \ref{fig:billiards_setup}.
In practice, we show that we can relax $z\in \{ 0, 1 \}^4$ into $a=[a_1, a_2, a_3, a_4]\in[0,1]^4$ and express $a$ as a function of $u_0$ (see Appendix \ref{subapp:billiard_wallhits_fn}). This allows us to simply define our \gls{SVGD} particles as $\theta = [u_0]^\top$, and optimize the task variable implicitly, as $a$ is a function of $u_0$ and optimizing $u_0$ will implicitly optimize $a$. To encourage diversity in the task plans, we employ $a$ along with $u_0$ within the kernel. We refer the reader to Appendix \ref{app:billiards_kernel} for details on how the kernel is defined for this problem.

Now we show that the task variable $z_{1:4}$ can be formulated as a soft function of the initial cue ball velocity $u_0=(v_x,v_y)$. To do so, we first simulate the cue ball's trajectory in a differentiable physics simulator, given the initial cue ball velocity $u_0=[v_x,v_y]$. We then use the rolled-out trajectory of the cue ball $x_{1:T}^\text{cue}$ to define a notion of distance at time $t$ between the cue ball and wall $i$ as follows.
\begin{equation}
	d_t^\text{wall-cue} = - \alpha\text{ SignedDistance}(x_t^\text{cue},\text{wall}_i) + \beta
\end{equation}
In the above, $\alpha>0,\beta\in\mathbb{R}$ are hyperparameters that can be tuned to scale and shift the resulting value. 

Then, we use $d_t^\text{wall-cue}$ to formulate the task variable $z_{1:4}$ as a soft function of $u_0$ as follows.
\begin{align}
    z_i &= \frac{1}{1+\exp{-d_\text{weighted}}}, \\
    d_\text{weighted} &= \sum_{t=1}^{t_c} \sigma_t d_t^\text{wall-cue}, \\
    \sigma_t &= \frac{\exp{d_t^\text{wall-cue}}}{\sum_{k=1}^{t_c}\exp{d_k^\text{wall-cue}}}
\end{align}
Although $z_i$ is not directly related to $u_0$, we note that both quantities are related implicitly since $d_t^\text{wall-cue}$ is a function of $x_t^\text{cue}$, which result from simulating the cube ball forward using $u_0$ as the initial ball velocity.

We note that the above formulation of $z_i$ gives us the binary behavior we want for the task variable, but in a relaxed way. That is, $z_i$ will either take on a value close to 1 or close to 0. Intuitively, the sigmoid function will assign a value close to $1$ to $z_i$ if the negative signed distance between the wall $i$ and the cue ball is large at some point in the cue ball's trajectory up until $t_c$, which is the time of contact with the target ball. This can only occur if the cue ball hits wall $i$ at some point in its trajectory before time $t_c$. Conversely, if the cue ball remains far from wall $i$ at every time step up to $t_c$, $z_i$ will correctly evaluate to a value close to 0. Note that $\alpha$ and $\beta$ can be tuned to get the behavior we want for $z_i$. 

Critically, we note that the above formulation is differentiable with respect to $u_0$. By relaxing $z_i$ using a sigmoid function over $d_\text{weighted}$ and using a differentiable physics simulator to roll out the cue ball's trajectory, we effectively construct $z_i$ as soft and differentiable functions of $u_0$. The differentiability of $z_i$ with respect to $u_0$ is important; because of the way we formulate the kernel (see Appendix \ref{app:billiards_kernel}), the repulsive force $\nabla_\theta k(\theta,\theta\prime)$ in the \gls{SVGD} update requires gradients of $z_i$ with respect to $u_0$ via the chain rule.

\figBilliard

\subsection{Kernel Definition}
\label{app:billiards_kernel}

In \gls{SVGD}, the kernel is used to compute a kernel-weighted sum of the gradient of the log posterior and to compute repulsive force. The kernel must be positive-definite.

We build upon the Radial Basis Function (RBF) kernel, a popular choice in the \gls{SVGD} literature, to design a positive definite kernel for $\theta=[v_x,v_y]^\top$. Whereas RBF kernels operate over $\theta$, we operate the RBF kernel over $[\theta,z_{1:4}(\theta)]^\top$, and tune separate kernel bandwidths for $v_{x,y}$ and $z_{1:4}(v_{x,y})$, which we denote as $s_{v_{x,y}}$ and $s_{z}$, respectively. This is to account for differences in their range. We use the median heuristic \cite{garreau2017large} to tune the kernel bandwidths.
\begin{equation}
    k(\theta, \theta^\prime) = \exp{ -\frac{\bigl\lVert \theta - \theta^\prime \bigr\rVert_2^2}{s_{v_{x, y}}^2} - \frac{ \bigl\lVert z_{1:4}(\theta) - z_{1:4}(\theta') \bigr\rVert_2^2 }{s_{z_{1:4}}^2} }
\end{equation}

\subsection{Loss Definitions}
\label{subapp:billiard_loss}
We define the target and aim loss for the billiards problem below. Given the final state at time $T$ of the target ball $x_T^\text{target}$, the position of the top pocket $g_\text{top}$, the position of the bottom pocket $g_\text{bottom}$, and the radius $R$ of the balls, the target loss is defined as the following. 
\begin{equation}
	L_\text{target} = \min(\lVert x_T^\text{target} - g_\text{top} \rVert_2^2, \lVert x_T^\text{target} - g_\text{bottom} \rVert_2^2)
\end{equation}

To define the aim loss, we first introduce the time-stamped aim loss $L_\text{aim}^{(t)}$ which is equal to the distance between the closest points on the surface of the target ball and the surface of the cue ball at time $t$:
\begin{equation}
	L_\text{aim}^{(t)} = \max(0, \abs{x_t^\text{target} - x_t^\text{cue}} - 2R)
\end{equation}

The aim loss $L_\text{aim}$ is the softmin-weighted sum of the time-stamped aim loss throughout the cue ball's time-discretized trajectory up until time $t_c$, which is the time of first contact with the target ball.
\begin{align}
    L_\text{aim} &= \sum_{t=1}^{t_c} \sigma_t L_\text{aim}^{(t)}, &
    \sigma_k &= \frac{\exp{-L_\text{aim}^{(k)}}}{\sum_{t=1}^{t_c} \exp{-L_\text{aim}^{(t)}}}
\end{align}

\section{Block-Pushing Problem Setup}
\label{app:pusher_setup}
\figPusher

\subsection{Graphical Overview}
Figure \ref{fig:blockpusher_setup} shows a graphical overview of the block-pushing problem environment, which we recreate in the Warp simulator. There are a total of four sides to push against: North, South, East, and West. Each of these constitute one symbolic action (e.g.,``push the North side"); hence, the action space is $|\mathcal{Z}|=4$. We represent each action numerically via one-hot encodings of size 4, by assigning each side to numbers 1-4 as shown in the figure.

\subsection{Formulation of the Posterior Distribution}
\label{app:pusher_loss}

Recall that the total cost function is defined:
\begin{equation}
    C(\theta) = \beta_\text{target}L_\text{target}(x_{KT}) + \beta_\text{traj}L_\text{traj}(g_{1:K}^{x,y})
\end{equation}

The target loss $L_\text{target}$ measures the proximity of the cube to the goal at the end of simulation, i.e. at timestep $KT$, and is defined as the following.
\begin{equation}
	L_\text{target} = \max\big(0, \abs{x_T^\text{block} - g_x} - t\big)^2 + \max\big(0, \abs{x_T^\text{block} - g_y} \big)^2
\end{equation}
The target loss forms our likelihood distribution $p(O=1|\theta)\propto\exp(-\beta_\text{target}L_\text{target})$.

The trajectory loss is a function of $g_k^{x,y}\forall k=1,...,K$ and penalizes the pursuit of indirect paths to the goal. We define it as:
\begin{align}
    L_\text{traj}(g_{1:K}^{x,y}) &= \sum_{k=1}^K \Bigl(D_\text{MH}(g_{k-1}^{x,y},g_k^{x,y}) \\
        &\qquad + D_\text{MH}(g_k^{x,y},g_K^{x,y}) \\
        &\qquad - D_\text{MH}(g_{k-1}^{x,y},g_K^{x,y})
    \Bigl)
\end{align}
where $D_\text{MH}(\cdot)$ is the Manhattan distance function and we use the convention $g_{x,y}^{(0)}=x_0^\text{cube}$ to denote the cube's starting state. As $L_\text{traj}$ is a function only of  $g_k^{x,y}$, it forms the prior $p(g_{1:K}^{x,y})$. In particular, we notice that
\begin{align}
    p\big(g^{x,y}_{1:K}\big) &= \prod_{k=1}^K p\big(g^{x,y}_{k} \mid g^{x,y}_{k-1}\big) \\
    p\big(g_{x,y}^{k} \mid g^{x,y}_{k-1}\big) &\propto \exp\Bigl( -D_\text{MH}\big(g^{x,y}_{k-1}, g_k^{x,y}\big) \\
        &\qquad - D_\text{MH}\big(g_k^{x,y}, g_K^{x,y}\big) \\
        &\qquad + D_\text{MH}\big(g_{k-1}^{x,y}, g_K^{x,y}\big) \Bigr)
\end{align}

The posterior distribution that we run inference over is the product of the likelihood and prior over the intermediate goals.
\begin{gather}
    p(\theta \mid O=1) \propto p(O=1|\theta)p(\theta) \nonumber \\ 
    \propto \exp{-\beta_\text{target} L_\text{target}(x_{KT})} \cdot \exp{ - \beta_\text{traj} L_\text{traj}(g_{1:K}^{x,y})}
\end{gather}

\subsection{Kernel Definition}
\label{app:pusher_kernel}

In \gls{SVGD}, the kernel is used to compute a kernel-weighted sum of the gradient of the log posterior and to compute repulsive force. The kernel must be positive-definite.

We employ the additive property of kernels to construct our positive-definite kernel. The kernel is a weighted sum of the RBF kernel over $\{g^{x,y}_k\}_{k=1}^K$, the RBF kernel over $\{\text{softmax}(a_k)\}_{k=1}^K$, and the von Mises kernel over $\{g_k^\phi\}_{k=1}^K$. The von Mises kernel is a positive definite kernel designed to handle angle wrap-around.

\begin{equation}
    \begin{aligned}
        K(\theta, \theta^\prime) &= \sum_{k=1}^K \Bigl[ \alpha_{g^{x,y}} K_\text{RBF}\Bigl( g^{x,y}_k, {g_{k}^{x,y}}^\prime \Bigr) \\
        &\qquad + \alpha_{z} K_\text{RBF}\Bigl(\softmax\bigl( a_k \bigr), \softmax\Bigl( a_k^\prime \Bigr) \Bigr) \\
        &\qquad + \alpha_{g^\phi} K_\text{VM}\Bigl( g^\phi_{k},  {g^\phi_{k}}^\prime \Bigr) \Bigr]
    \end{aligned}
\end{equation}

\subsection{Further Results}
Figure~\ref{fig:pushing_sols} shows sample solutions obtained from running STAMP on the block pushing problem.

\section{Pick-and-Place Problem Setup}
\label{section: pick-place-appdx}

\subsection{Graphical Overview}

\begin{figure}[!t]
    \centering
    \vspace{0.2cm}
    {\includegraphics[width=.75\linewidth]{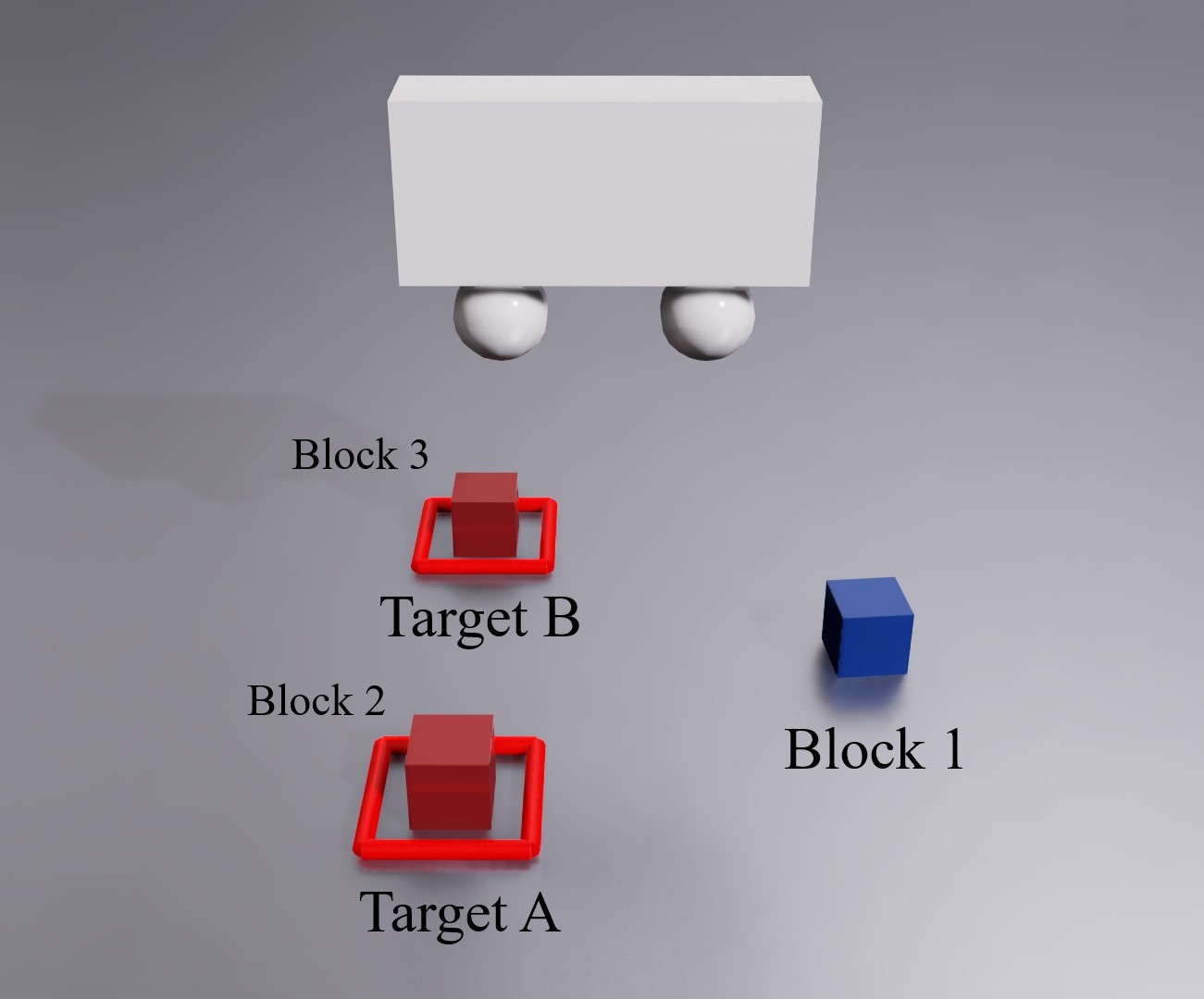}}
    \caption{Graphical depiction of the pick-and-place problem environment.}
    \label{fig:bs_setup}
\end{figure}

The goal of the pick-and-place problem is to place block 1 into either of the two targets, $\mathcal{A,B}$, without placing it on top of blocks 2 or 3 which occlude the targets. Correct solutions to this problem involve displacing either block 2 or 3 from the targets and placing block 1 into the empty target. Figure \ref{fig:bs_setup} shows a graphical depiction of the problem environment, created and rendered in the Warp simulator. The gripper (or robot end effector) is used to manipulate the blocks.

\subsection{Precondition and Postcondition Loss Definitions}
\label{subsection:prepost_defn}

Symbolic actions for the pick-place environment include $\mathcal{Z}=\{\texttt{pick(cube), \texttt{place(cube)}}\}$ $\forall \texttt{cube}\in\mathcal{C}$ (where $\mathcal{C}=\{\texttt{cube1}, \texttt{cube3}, \texttt{cube3}\}$).

\begin{figure*}[t!]
    \centering
    \subfloat[Particles at initialization]{\includegraphics[width=.25\linewidth]{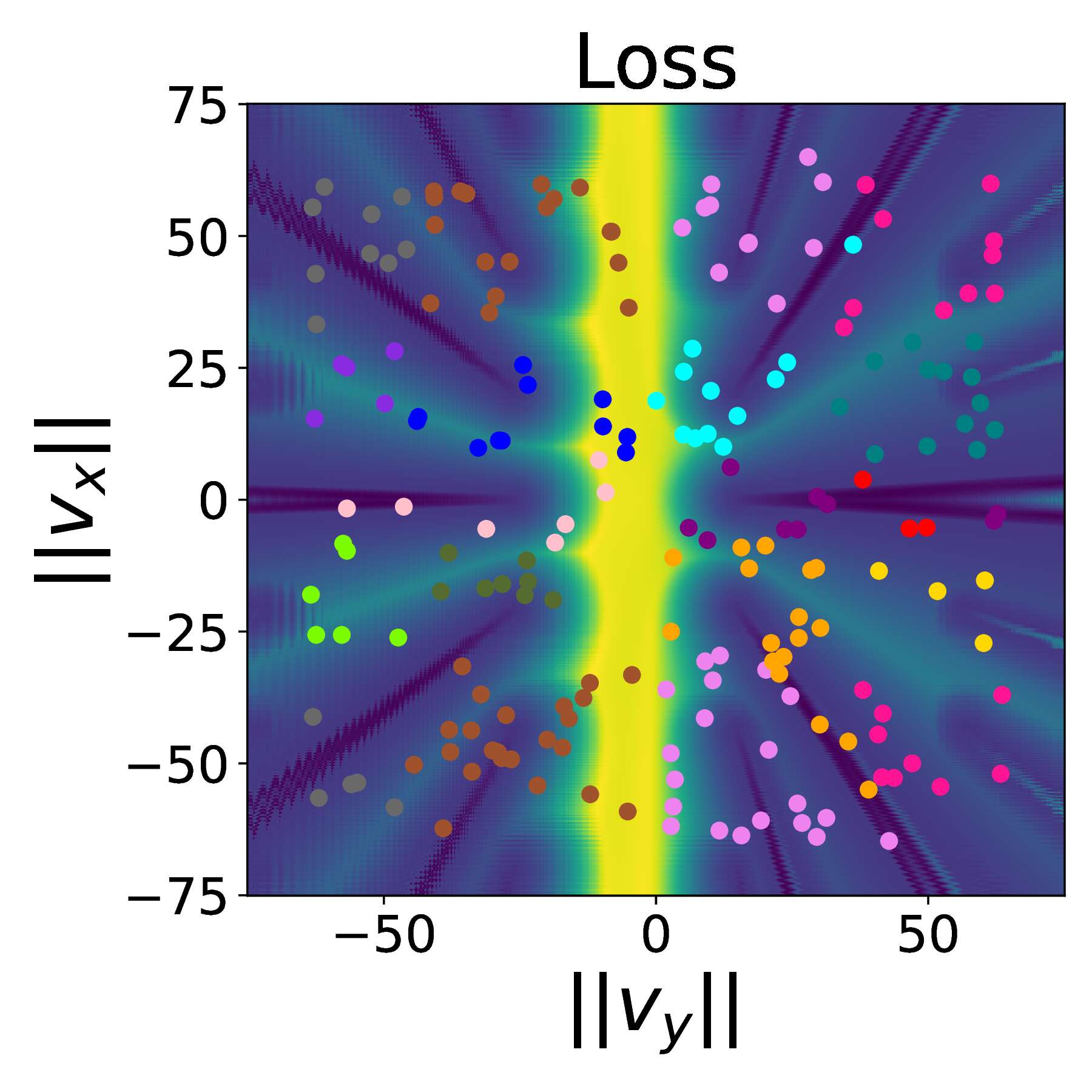}}
    \qquad
    \subfloat[Particles after SVGD inference]{\includegraphics[width=.25\linewidth]{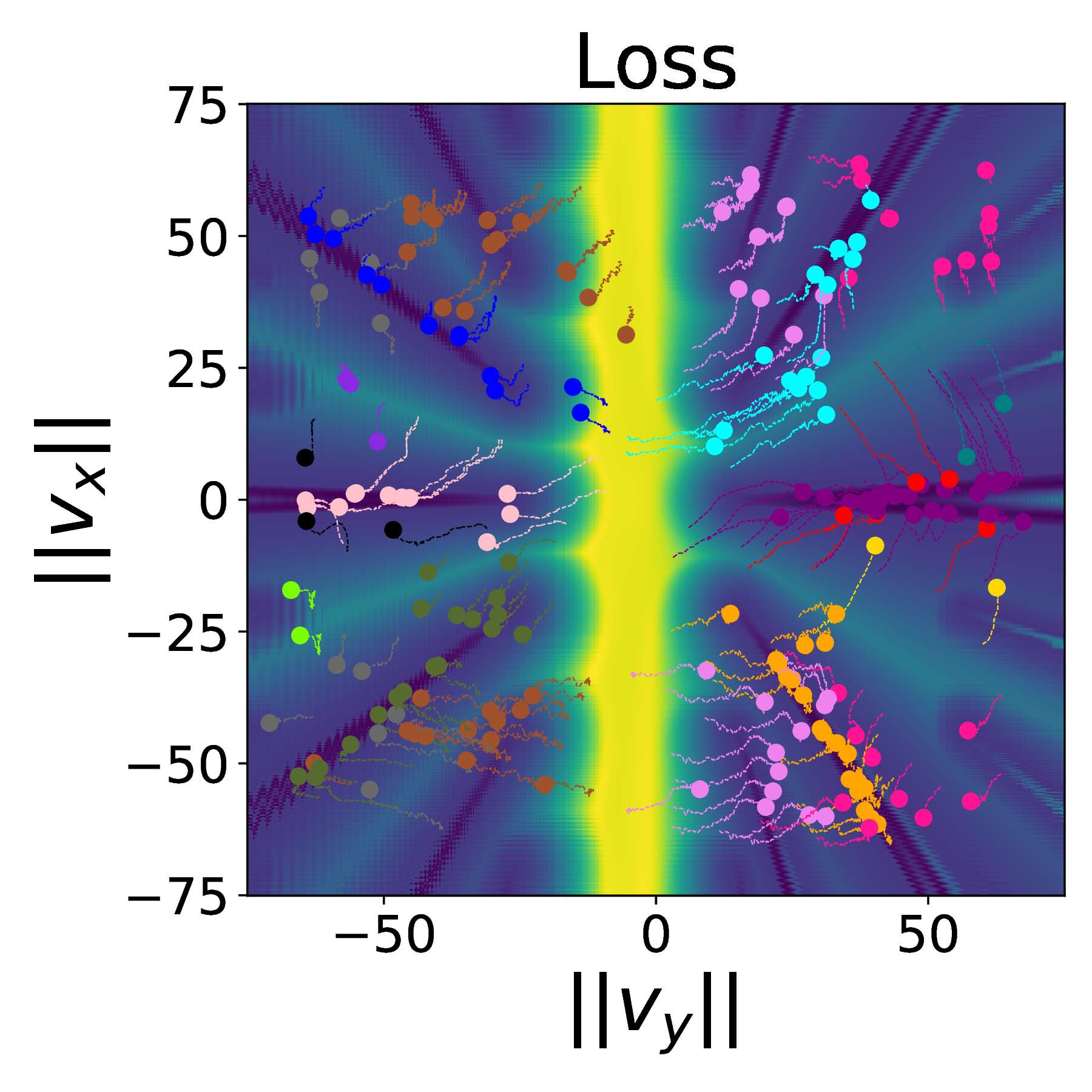}}
    \qquad
    \subfloat[Particles after SGD refinement]{\includegraphics[width=.25\linewidth]{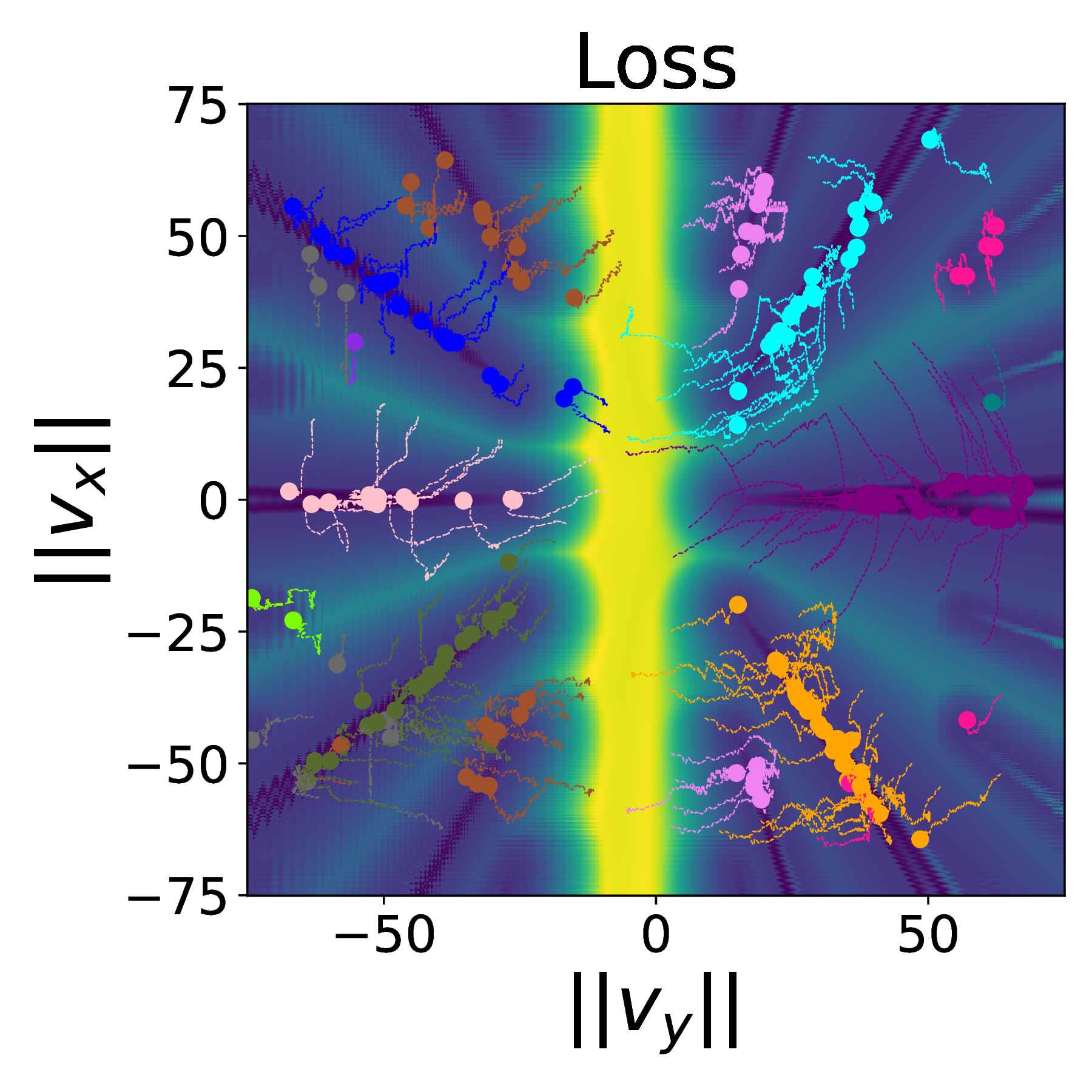}}
    \caption{From left to right: evolution of the distribution of plan parameters $\theta$. On the left, $\theta$ are initialized uniformly, while in the middle, they converge to the full likelihood distribution post-SVGD. On the right, the particles collapse to the optima post-SGD. The particles' colors indicate their mode (wall hits).}
    \label{fig:optim_anim}
\end{figure*}

Preconditions for $\texttt{pick(cube)}$ are $\texttt{pre\big(pick(cube)\big)} = \{\texttt{eeFree}, \texttt{canGrasp(cube,}g\texttt{)}\}$ -- that is, the end effector is free and the at goal $g$ the cube is graspable. Postconditions for $\texttt{pick(cube)}$ are $\texttt{post\big(pick(cube)\big)} = \{\texttt{eeHolding(cube)}\}$ - that is, the end effector is holding the cube. Meanwhile, $\texttt{pre\big(place(cube)\big)} = \{\texttt{eeHolding(cube)}, \texttt{canPlace(cube,}g\texttt{)}\}$; that is, the end effector is holding the cube and the cube can be placed at goal $\texttt{g}$. Finally, $\texttt{post\big(place(cube)\big)} = \{\texttt{eeFree}\}$. 

In general, precondition losses are defined w.r.t. the initial states $x_{t_0}$ and pre-emtively sampled goal state $g$ (from the particle), while postcondition losses are defined w.r.t the final states $x_{t_T}$. Pre/postcondition losses can be defined differentiably, in many cases, through the use of squared distance functions; see the following subsections for details.

\subsubsection{Loss for Condition \texttt{eeFree} $\rightarrow$ $L_\texttt{eeFree}$}

\texttt{eeFree} is the condition in which the robot end effector is not holding an object. The associated loss function $L_\texttt{eeFree}$ is minimized (evaluates to $\approx 0$) if $\texttt{eeFree=True}$.

We first define two distance metrics, $d_{\texttt{ee-}c}$ and $d_\texttt{grippers}$, which are shifted forms of the squared distance function. $d_{\texttt{ee-}c}^{(t)}$ measures the distance between the end effector and cube $c$ at time $t$, while $d_\texttt{grippers}^{(t)}$ measures the distance between the grippers at time $t$. These can be expressed as:
\begin{align}
    d_{\texttt{ee-}c}^{(t)} &= \text{dist}(x_t^\texttt{ee},x_t^c) \\
    d_\texttt{grippers}^{(t)} &= \text{dist}(x_t^\texttt{left gripper}, x_t^\texttt{right gripper})
\end{align}
where dist($\cdot$) represents the distance function. When the end effector is holding onto cube $c$, the grippers are closed and the distance between the end effector and the cube is minimized; thus, $d_{\texttt{ee-}c}, d_\texttt{grippers}\approx0$. Meanwhile, when $\texttt{eeFree=True}$, $d_{\texttt{ee-}c}, d_\texttt{grippers}\gg0$.

We define the $\texttt{eeFree}$ loss $L_\texttt{eeFree}$ as:
\begin{equation}
    L_\texttt{eeFree}(x_t) = \text{max}_{c}\\\big(L_{\texttt{eeFree},c}(x_t)\big),
\end{equation}
where
\begin{equation}
    L_{\texttt{eeFree},c}(x_t) = -\log\Bigg(1-\exp(-d_{\texttt{ee-}c}^{(t)} - d_\texttt{grippers}^{(t)})\Bigg).
\end{equation}
$L_{\texttt{eeFree},c}>0$ grows large as both $d_{\texttt{ee-}c}, d_\texttt{grippers}\rightarrow0$ (i.e., as the probability that the end effector is holding $c$ increases). Since
$L_\texttt{eeFree}(x_t)>0$ equals $L_{\texttt{eeFree},c}>0$ for the cube $c$ with the highest probability of being `held' by the gripper, if $L_\texttt{eeFree}(x_t)\rightarrow0$, there is a high likelihood that $\texttt{eeFree=True}$.

\subsubsection{Loss for Condition \texttt{eeHolding}(c) $\rightarrow$ $L_{\texttt{eeHolding}(c)}$}

The loss function for $\texttt{eeHolding}(c)$ is simply the sum of the distance functions between the end effector and cube $c$, as well as between the grippers:
\begin{equation}
    L_{\texttt{eeHolding}(c)}(x_t) = d_{\texttt{ee-}c}^{(t)} + d_\texttt{grippers}^{(t)}
\end{equation}
$L_{\texttt{eeHolding}(c)}\geq0$ is optimal when $d_{\texttt{ee-}c}, d_\texttt{grippers}\approx0$, which is true when the end effector is holding cube $c$.

\subsubsection{Loss for Condition \texttt{canGrasp}(c,g) $\rightarrow$ $L_{\texttt{canGrasp}(c,g)}$}

$\texttt{canGrasp}(c,g)=\texttt{True}$ when the end effector can grasp cube $c$, and it is $\texttt{False}$ otherwise. There are two conditions that must be satisfied for $\texttt{canGrasp}(c,g)=\texttt{True}$: first, cube $c$ must be graspable (e.g., no other cubes are stacked on top); second, the cube is graspable at the pre-emptively sampled goal pose $g$.

$L_{\texttt{canGrasp}(c,g)}$ is the sum of losses that express these two conditions:
\begin{equation}
    L_{\texttt{canGrasp}(c,g)}(x_t,g) = L_{\texttt{cubeFree}(c,g)}(x_t) + d_{c-g}^{(t)}(x_t,g), 
\end{equation}
where $L_{\texttt{cubeFree}(c)}\geq0$ is optimal ($=0$) when no cubes are stacked above $c$ and is otherwise equal to the squared vertical distance, distVertical($\cdot$), between $c$ and the cube stacked directly above it.
\begin{align}
    L_{\texttt{cubeFree}(c)} =
    \begin{cases}
        0 \text{ if }\text{height}(x_t^c) \geq \text{height}(x_t^j) \forall j \in\mathcal{C}, j\neq c, \\
        \min_{j\in\mathcal{C}, j\neq c}\big(\text{distVertical}(x_t^c,x_t^j)\big) \text{ otherwise}
    \end{cases}
\end{align}
Meanwhile, $d_{c-g}^{(t)}$ measures the squared distance between cube $c$ and the sampled goal pose $g$.
\begin{equation}
    d_{c-g}^{(t)} = \text{dist}(g,x_t^c)
\end{equation}

\subsubsection{Loss for Condition \texttt{canPlace}(c,g) $\rightarrow$ $L_{\texttt{canPlace}(c,g)}$}

$\texttt{canPlace}(c,g)=\texttt{True}$ when two conditions are satisfied: first, the pre-emptively sampled goal pose, $g$, where the cube $c$ will be placed, must be close to the ground; second, $g$ should not coincide with where other cubes $j\in\mathcal{C}$ are located. We express $L_{\texttt{canPlace}(c,g)}$ as the following sum:
\begin{equation}
    L_{\texttt{canPlace}(c,g)} = \text{dist}(g,\texttt{ground}) + \sum_{j\in\mathcal{C},j\neq c}\exp\big(-\text{dist}(x_t^j,g)\big).
\end{equation}
The first term in the above is minimal when $g$ is close to the ground, while the second term is minimal when all cubes $j\in\mathcal{C}$ are far from $g$.

\subsection{Definition of $\omega_c$}
\label{app:holding_like_def}

We define $\omega_c$ as:
\begin{equation}
    \omega_c(x_t) = -\log\Big(1-\exp(d_{\texttt{ee-}c}^{(t)} - d_\texttt{grippers}^{(t)})\Big),
\end{equation}
where $d_{\texttt{ee-}c}^{(t)}$ and $d_\texttt{grippers}^{(t)}$ are as defined in section \ref{subsection:prepost_defn}.

\subsection{Target Loss Definition}
\label{app:pick_place_target_loss}
\begin{gather}
    L_{\text{target,}c}(x_{t_T}) = 
        \begin{cases}
            \text{dist}(x_{t_T}^c,\mathcal{R}) \text{ if }c=\{\texttt{cube1,cube2}\}, \\
            \begin{aligned}
                &\mathds{1} \{\zeta_{\mathcal{A}} \geq \zeta_{\mathcal{B}}\} \cdot \text{dist}(x_{t_T}^\texttt{cube1}, \mathcal{A}) + \\
                &\mathds{1} \{\zeta_{\mathcal{A}} < \zeta_{\mathcal{B}}\} \cdot \text{dist}(x_{t_T}^\texttt{cube1}, \mathcal{B}) \text{ otherwise, }
            \end{aligned}
        \end{cases}
\end{gather}
where $\mathcal{R}=\lnot(\mathcal{A}\land\mathcal{B})$, $x_{t}^c$ is cube $c$'s position at time $t$, $\text{dist}(\cdot)$ is a differentiable distance function (e.g. squared distance), and $\zeta_\mathcal{A} = \text{dist}(\texttt{cube2},\mathcal{A})$, $\zeta_\mathcal{B}=\text{dist}(\texttt{cube3},\mathcal{B})$ differentiably mimic indicator functions which output a large number when $\mathcal{A,B}$ are free and 0 if they are occluded by \texttt{cube1, cube2}, respectively.

\figPusherSolutions

\subsection{Kernel Definition}

As the pick-and-place particle definition is nearly identical to that of the block pushing problem, the kernel is defined similarly. The kernel for the block pushing problem environment is defined in Section \ref{app:pusher_kernel}.

\subsection{Exploding Gradient Issues}
\label{subsection:block_opt}
A potential limitation of employing differentiable simulators is that gradients can explode as the time horizon of the simulation increases. This can be a problem when trying to run \gls{STAMP} on long-horizon sequential manipulation problems, such as pick-and-place.

Here, we propose a modification to \gls{STAMP} that mitigates gradient explosion due to long horizon simulation. The key insight we leverage is that the cost function, as defined in equation \eqref{eq: cost_bs}, is expressed entirely as the sum of individual loss terms which only depend on trajectories within each of the $K$ action sequences; i.e., it is of the form:
\begin{equation}
    C(\theta) = \sum_{k=1}^{K} C_k(a_k,g_k,x_{(k-1)T+1:kT})
\end{equation}

As the loss functions within the sums only depend on short trajectories (e.g., $x_{(k-1)T+1:kT}$ as opposed to $x_{1:KT}$) and the task variable $a_k$, we can split the optimization into smaller chunks by defining the posterior using the `inner' cost $C_k$, which prevents the need for taking gradients over the entire trajectory $x_{1:KT}$ and mitigates the danger of gradient explosion during optimization.

\section{Additional Results}

\subsection{Block Pushing Results}
Figure \ref{fig:pushing_sols} shows samples of solutions obtained from running \gls{STAMP} on the block pushing experiment.

\subsection{SVGD-SGD Plan Refinement}
\label{ref:svgd_svgd_visualization}

Figure \ref{fig:optim_anim} shows how the particles move through the loss landscape of the billiards problem.
Figure \ref{fig:loss_evolution} shows how the loss evolves over various iterations of \gls{STAMP} for the billiards and block-pushing problems.


\section{Implementation of Baselines} \label{app:baselines}

For both baselines, the  task plan is the walls to hit for the billiards problem and the sequence of sides to push from for the block pushing problem.

\subsection{PDDLStream}

PDDLStream combines search-based classical planners with \textit{streams}, which construct optimistic objects for the task planner to form an optimistic plan and then conditionally sample continuous values to determine whether these optimistic objects can be satisfied \cite{Garrett2020-ts}. 
Since it is possible to create an infinite amount of streams and thus optimistic objects, the number of stream evaluations required to satisfy a plan is limited and incremented iteratively. As a result, PDDLStream will always return the first-found plan with the smallest possible sequence of actions in our evaluation problems, resulting in no diversity. 

To generate different solutions, we force PDDLStream to select different task plans based on some preliminary work in \url{https://github.com/caelan/pddlstream/tree/diverse}. 
For the billiards problem, we can reduce the task planning problem by enumerating all possible wall hits and denoting each sequence as a single task, since the motion planning stream is only needed once for any sequence of wall hits. Then, we select a task by uniform sampling and try to satisfy it by sampling initial velocities that match the desired wall hits using \gls{SGD} and Warp. For the block-pushing problem, we use a top-k or diverse PDDL planner to generate multiple optimistic plans of varying complexity. We then construct the problem environment with the same Warp environment as STAMP and use the pretrained DMPs for the motion planning streams. Note that the distribution of solutions is highly dependent on the time allocated for sampling streams. Of the feasible candidate plans, PDDLStream randomly selects a solution. 

\subsection{Logic-Geometric Programming}

Diverse \gls{LGP}~\cite{Ortiz-Haro_Karpas_Toussaint_Katz_2022} uses a two-stage optimization approach by first formulating the problem on a high, task-based level (as an SAS\textsuperscript{+}~\cite{backstrom1995complexity} task).
Subsequently, a geometric (motion planning) problem is solved conditioned on the logical plan.
That is, the performed motion is required to fulfill the logical plan.
A logical plan is called \emph{geometrically infeasible} if there is no motion plan fulfilling it.
Diverse \gls{LGP} speeds up exploration by eagerly forbidding geometrically infeasible plan prefixes (for instance, if moving through a wall is geometrically infeasible, no sequence starting with moving through a wall has to be tested for geometric feasibility).

We use the first iteration of LAMA~\cite{richter2010lama} (implemented as part of the Fast Downward framework~\cite{Helmert_2006}) as suggested in \cite{Ortiz-Haro_Karpas_Toussaint_Katz_2022} and use \gls{SGD} for motion planning. 
To enforce diversity in the logical planner, we iteratively generate \num{25} logical plans by blocking ones we already found.
As Diverse \gls{LGP} finds exactly one solution to the \gls{TAMP} problem per run, we invoke it as many times as \gls{STAMP} found solutions to get the same sample size.

For both billiards and block-pushing, we use the \gls{SGD} component described in Section~\ref{sec:stamp_algo_description}.
For billiards, we additionally add a loss term ensuring that the required walls are hit.
For pusher, it suffices to fix $z_1, \dots, z_K$ in $\theta$.
That is, excluding them from the \gls{SGD} update.
\end{document}

%% file: figures.tex
\newcommand{\figMixtureOfGaussians}{
    \begin{figure}
        \centering
        \vspace{0.2cm}
        \includegraphics[width=\linewidth]{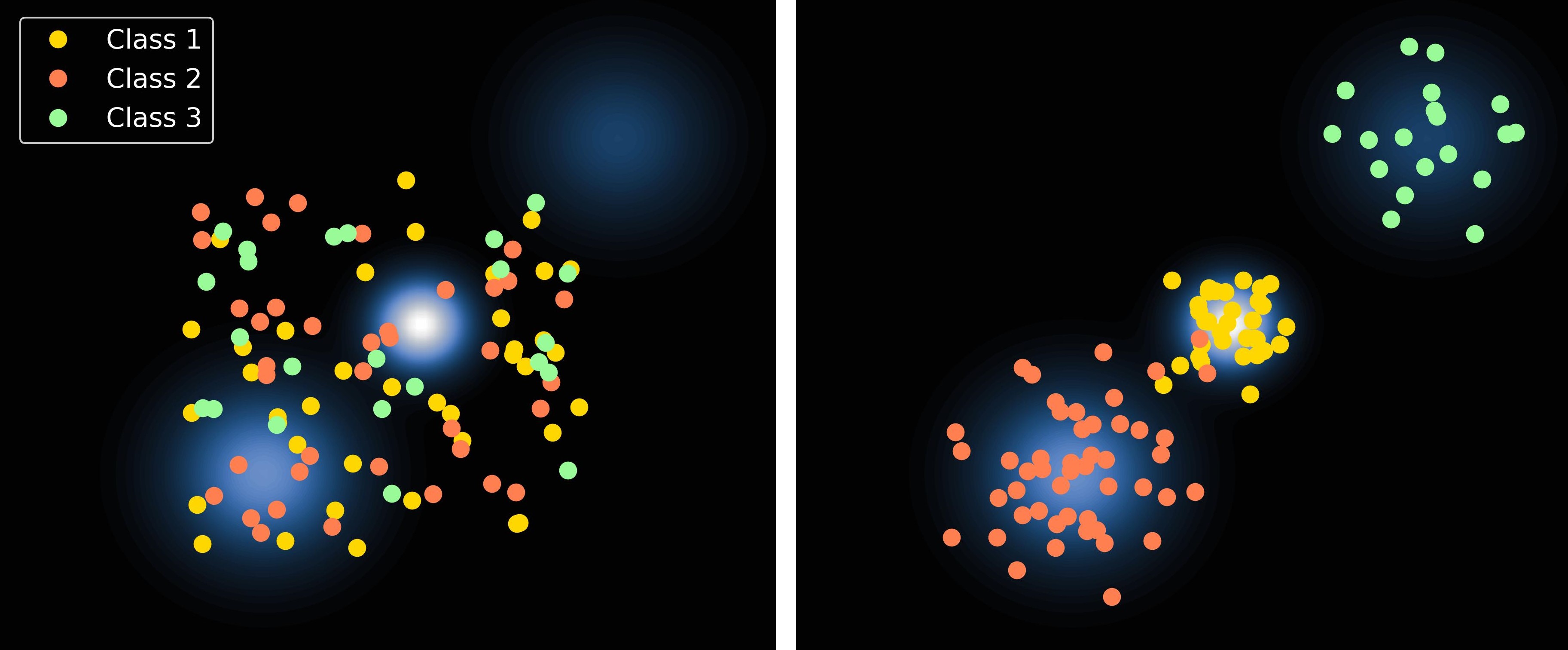}
        \setlength{\abovecaptionskip}{-0.4cm} 
        \caption{Fitting a Gaussian mixture with discrete-and-continuous SVGD. Left: random initialization of particles. Right: particles after 500 updates. Note that same-color particles have been assigned to the same mode.} 
        \vspace{-0.45cm}
        \label{fig:svgd_gm}
    \end{figure}
}

\newcommand{\figStampPipeline}{
    \begin{figure*}
        \centering
        \vspace{0.2cm}
        \setlength{\abovecaptionskip}{-0.2cm} 
        \includegraphics[width=\linewidth]{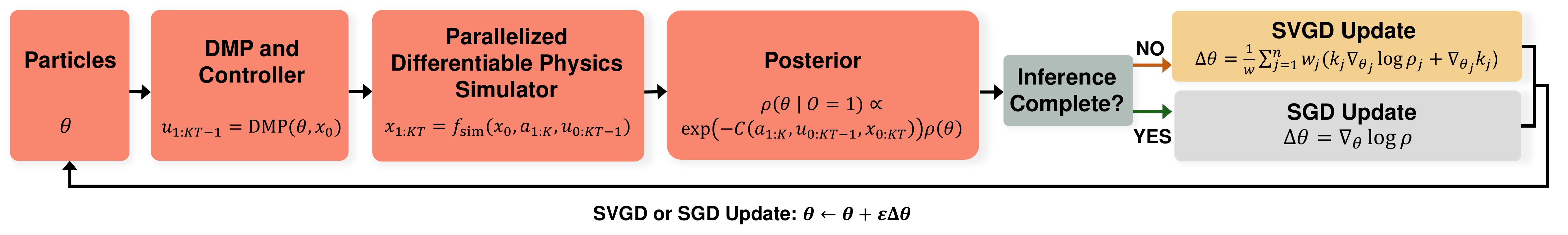}
        \caption{Overview of STAMP algorithm pipeline. The particles $\theta$ represent the task and motion plan $\theta=[a_{1:K}, u_{0:KT-1}]$, or the task plan and a parametrization of the motion plan $\theta=[a_{1:K}, g_{1:K}]$ if Dynamic Movement Primitives dependent on goals $g_{1:K}$ are used (see section \ref{sec:dmp_for_stamp}). The resulting controls are passed through a differentiable physics simulator, and particles are updated in parallel using two phases of optimization: an inference phase of SVGD that balances loss minimization with diversification of particles followed by a finetuning phase of SGD so that particles can reach the local optima.}
        \vspace{-0.4cm} 
        \label{fig:pipeline} 
    \end{figure*}
}

\newcommand{\figBilliardSolutions}{

   \begin{figure}
        \vspace{0.2cm}
       \includegraphics[width=\linewidth]{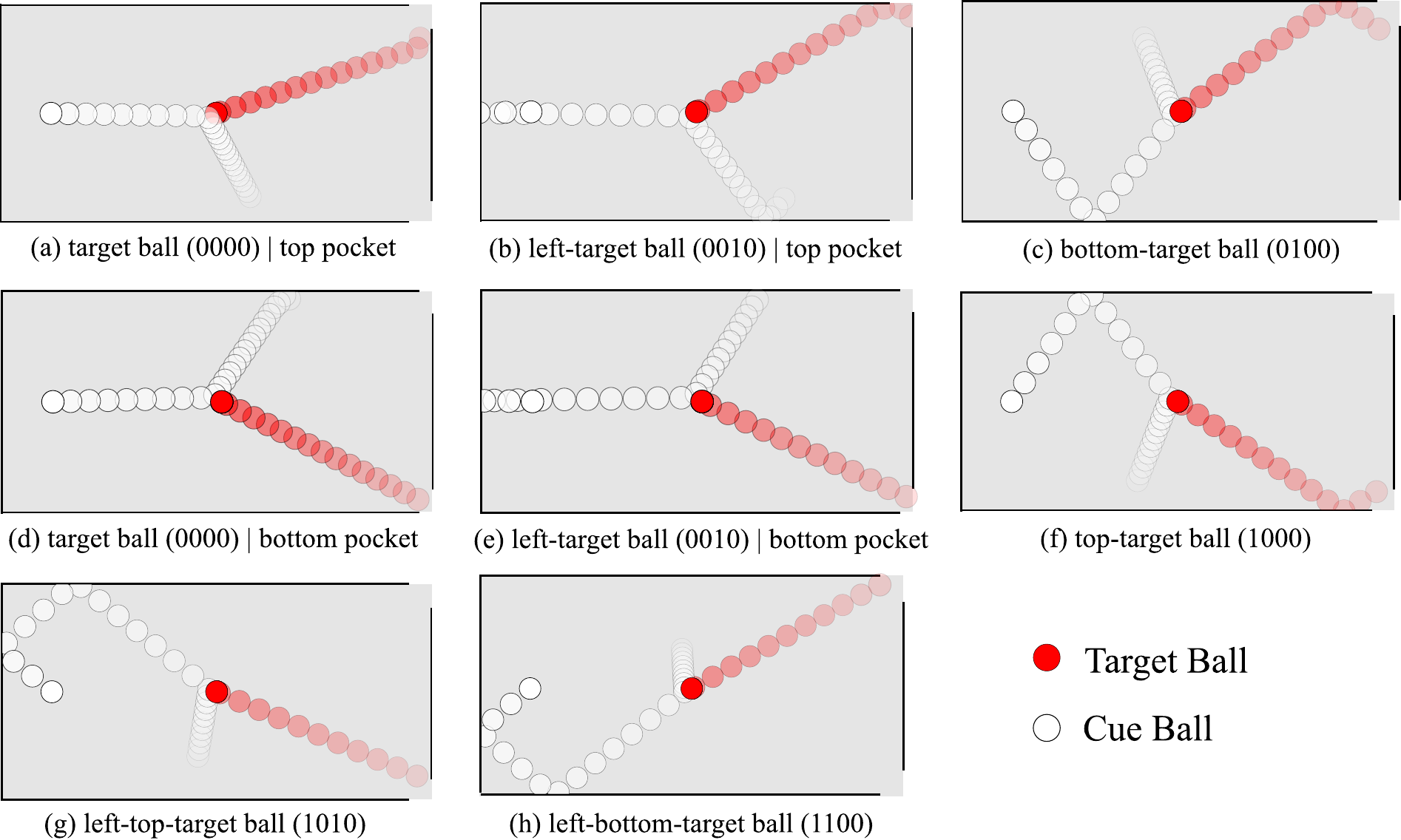}
       \setlength{\abovecaptionskip}{-0.4cm} 
       \caption{Sample solutions STAMP found for the billiards problem. Subcaptions indicate which walls the cue ball hits before hitting the target ball. The $1^\text{st}$/$2^\text{nd}$/$3^\text{rd}$/$4^\text{th}$ digit correspond to hitting the top/bottom/left/right wall. A 1 means the wall is hit during the shot; a 0 means that it is not. The caption also indicates in which pocket the target ball is shot.}
       \vspace{-0.6cm} 
       \label{fig:billiards_sols}
   \end{figure}
}

\newcommand{\figPusherSolutions}{
    \begin{figure*}[t!]
        \setlength{\abovecaptionskip}{5pt}
        \centering
        \subfloat[Push on north side.\label{fig:north_push}]{\includegraphics[width=\linewidth]{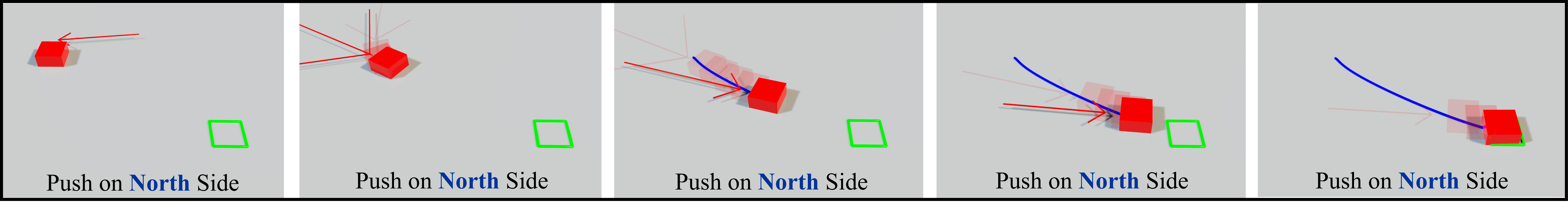}}
        \\[.2cm]
        \subfloat[Push on north side, followed by push on south side.\label{fig:bottom_top}]{\includegraphics[width=\linewidth]{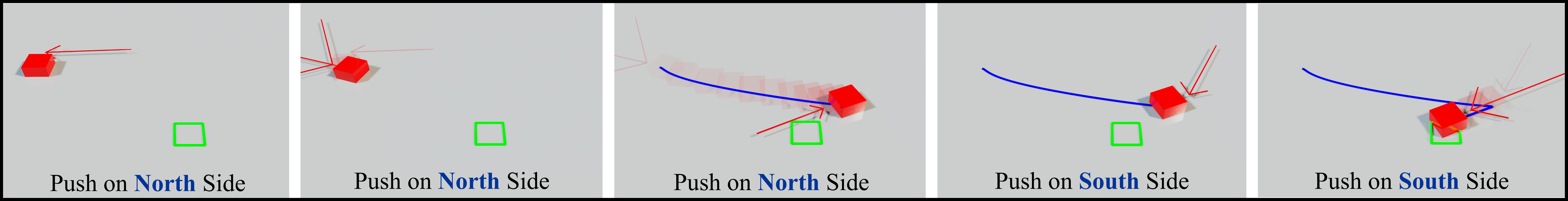}}
        \caption{Sample solutions obtained from running STAMP on the block pushing problem. The red arrow is proportional to the magnitude and direction of the pushing force, the blue line shows the block's trajectory, and the green box shows the goal region. Note, the notions of `north', `south', `east' and `west' are relative to the cube's frame of reference (see Figure \ref{fig:blockpusher_setup}).}
        \label{fig:pushing_sols}
    \end{figure*}
}

\newcommand{\figHists}{
    \begin{figure*}
        \subfloat[STAMP found \num{104} solutions in \SI{6.94 \pm 0.06}{\second}. Diverse LGP and PDDLStream are invoked \num{104} times. \label{fig:histo_billiards}]{\includegraphics[width=.34\linewidth]{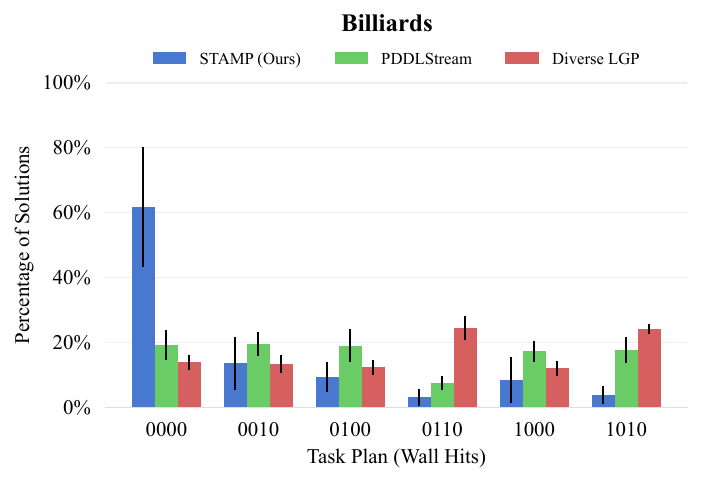}}
        \hfill
        \subfloat[STAMP found \num{825} solutions in \SI{12.92 \pm 0.04}{\second}. Diverse LGP and PDDLStream are invoked \num{150} times. \label{fig:histo_pusher}]{\includegraphics[width=.6\linewidth]{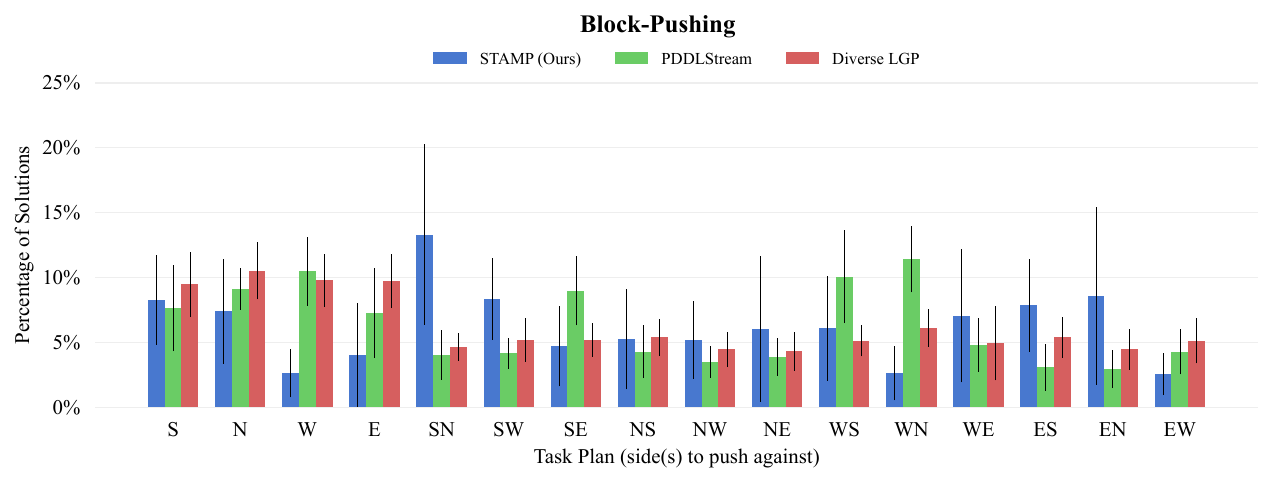}}
        \vspace{-0.4cm} 
        \caption{Distribution of plans found by STAMP (our method), Diverse LGP, and PDDLStream. Results are averages across \num{10} seeds; error bars represents standard deviation. Solutions for the remaining modes (not shown) were not found by any method.}
        \vspace{-0.5cm} 
    \end{figure*}
}

\newcommand{\figBilliard}{
    \begin{figure}
        \centering
        \vspace{0.2cm}
        \includegraphics[width=0.8\linewidth]{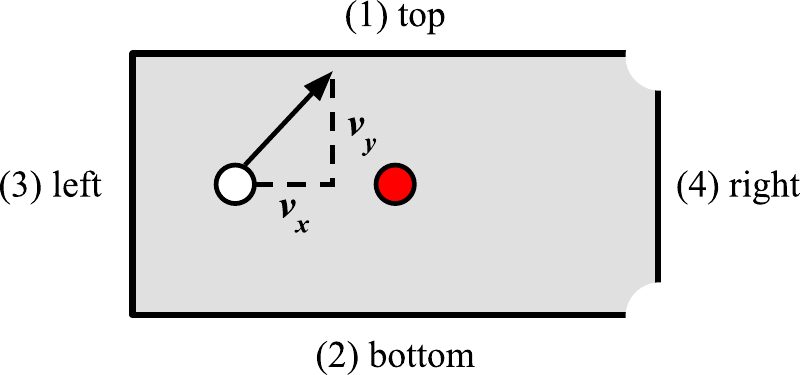}
        \caption{Graphical depiction of the billiards problem 
        environment. The goal is to shoot the target ball (red) into one of the pockets on the right by finding an initial velocity for the cue ball (white). The task plan is give by the wall(s) the cue ball should hit before hitting the target ball.}
        \label{fig:billiards_setup}
    \end{figure}
}

\newcommand{\figPusher}{
    \begin{figure}
        \centering
        \vspace{0.2cm}
        \includegraphics[width=.6\linewidth]{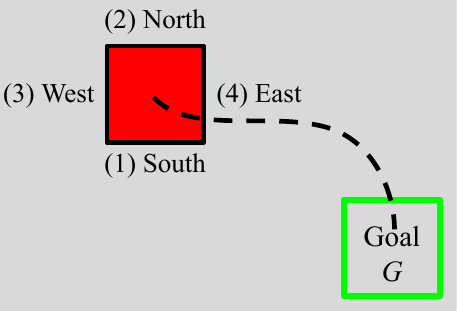}
        \caption{Graphical depiction of the block-pushing problem environment. The goal is to push the red block into the green goal region by repeatedly applying pushes to either north, south, west, or east of the red block, where these directions are defined relative to the \textit{cube's frame of reference}. The task plan is given by the side(s) to push from.}
        \label{fig:blockpusher_setup}
    \end{figure}
}

\newcommand{\tabRuntimeComparison}{
    \begin{threeparttable}
        \scriptsize
        \caption{STAMP's Runtime vs.\ Baselines}
        \label{tab:runtime_baselines}
        \begin{tabular}{c | l S[table-format = 2.4(1.4)] S[table-format = 3.2(1.2)]}
            \toprule
                & \textbf{Method} & {\textbf{Time / Solution} (\si{\second})} & {\textbf{Runtime} (\si{\second})} \\ \midrule
            \multirow{3}{*}{\rotatebox{90}{\textbf{Billiards}}}
                & Diverse LGP     & 7.5 \pm 0.9                               & 7.5 \pm 0.9                      \\
                & PDDLStream      & 5.7 \pm 0.8                               & 5.7 \pm 0.8                      \\
                & STAMP (Ours)    & 0.068 \pm 0.007                           & 6.94 \pm 0.06                    \\ \midrule
            \multirow{3}{*}{\rotatebox{90}{\textbf{Pusher}}}
                & Diverse LGP     & 22.6 \pm 7.6                              & 22.6 \pm 7.6                     \\
                & PDDLStream      & 22.2 \pm 0.1                              & 110.9 \pm 0.7                    \\
                & STAMP (Ours)    & 0.0157 \pm 0.0004                         & 12.92 \pm 0.04                   \\ \bottomrule
        \end{tabular}
        \begin{tablenotes}[para, flushleft]
            \scriptsize
            \item Averaged across \num{10} seeds; \num{1100} particles; max.\ \num{2} pushes.
        \end{tablenotes}
    \end{threeparttable}
}

\newcommand{\tabStampRuntimeParticles}{
    \begin{threeparttable}
        \scriptsize
        \caption{STAMP's Runtime varying the Number of Particles}
        \label{tab:runtime_num_particles}
        \begin{tabular}{S[table-format = 4.0(0.0)] S[table-format = 1.2(1.2)] S[table-format = 2.2(1.2)]}
            \toprule
            \textbf{\#Particles} & {\textbf{Billiards} (\si{\second})} & {\textbf{Pusher} (\si{\second})} \\ \midrule
            700                  & 6.30 \pm 0.07                       &  12.53 \pm 0.05                  \\
            900                  & 6.55 \pm 0.04                       &  12.74 \pm 0.03                  \\
            1100                 & 6.94 \pm 0.06                       &  12.92 \pm 0.04                  \\
            1300                 & 7.71 \pm 0.12                       &  13.22 \pm 0.06                  \\ \bottomrule
        \end{tabular}
        \begin{tablenotes}[para, flushleft]
            \scriptsize
            \item Averaged across \num{10} seeds; max.\ \num{2} pushes.
        \end{tablenotes}
    \end{threeparttable}
}

\newcommand{\tabStampRuntimeDimensionality}{
    \begin{threeparttable}
        \scriptsize
        \caption{STAMP's Runtime varying the Particle Dimensionality (Pusher)}
        \label{tab:runtime_dimensions}
        \begin{tabular}{S[table-format = 1.0(0.0)] S[table-format = 2.0(0.0)] S[table-format = 2.2(1.2)]}
            \toprule
            \textbf{Max.\ \#Pushes} & \textbf{Dim.} & {\textbf{Runtime} (\si{\second})} \\ \midrule    
            2                       & 16            & 12.92 \pm 0.04                    \\
            3                       & 24            & 12.93 \pm 0.04                    \\
            4                       & 32            & 13.26 \pm 0.03                    \\
            5                       & 40            & 13.57 \pm 0.03                    \\ \bottomrule
        \end{tabular}
        \begin{tablenotes}[para, flushleft]
            \scriptsize
            \item Averaged across \num{10} seeds; \num{1100} particles.
        \end{tablenotes}
    \end{threeparttable}
}

%% file: algo.tex
\SetKwComment{Comment}{$\triangleright$\ }{}
\begin{algorithm}
    \setstretch{1.2}
    \DontPrintSemicolon
    \textbf{let}: \textit{step size} = $\epsilon$, \textit{phase = SVGD} \;
    \textbf{initialize}: $n$ particles (candidate task and motion plans) $\{\theta_i\}_{i=1}^n$  randomly, where $\theta_i=[a_{1:K},u_{0:KT-1}]_i$ \;
    \While{not converged}{
        \Comment{simulate plans given by each particle in parallel}
        $\displaystyle [x_{1:KT}]_i = f_\text{sim}\big(x_0, \theta_i)$ \;  
        \BlankLine
        \Comment{compute posterior for all particles in parallel}
        $\displaystyle \rho(\theta_i | O\!=\!1)\!\propto \!\exp\!\{-C(\theta_i, [x_{0:KT}]_{i})\}\rho(\theta_i)$\!\!\!\!\;
        \BlankLine
        \Comment{compute update for all particles in parallel}
        \If{\text{phase = SVGD}}{
            $\displaystyle
                \begin{aligned} %
                    \Delta\theta_{i} &= \frac{1}{w}\sum_{j=1}^{n} w_j \bigl[\nabla_{\theta_j}\!\log \rho(\theta_j | O=1)k_{ji} + \nabla_{\theta_j} k_{ji} \bigr]  
                \end{aligned}
            $
            \If{\text{inference converged}}{
                \textit{phase = SGD}    
            }
        }
        \ElseIf{\text{phase = SGD}}{
            $\displaystyle \Delta\theta_{i} = \nabla_{\theta_i} \log \rho(\theta_i \mid O = 1)$ \;
            \BlankLine
        }
        $\theta_i \leftarrow \theta_i + \epsilon\Delta\theta_{i}$ \text{in parallel} \;
    }
    \caption{Stein TAMP (STAMP)}
    \label{algo:svgd-sgd}
\end{algorithm}